\documentclass[10pt,twocolumn,letterpaper]{article}

\usepackage[pagenumbers]{cvpr}
\usepackage{capt-of}

\usepackage{amsmath}
\usepackage{booktabs}
\usepackage{tabularx}
\newcommand{\method}{FoundationCov}
\usepackage{xcolor}

\renewcommand{\paragraph}[1]{\vspace{.5em}\noindent\textbf{#1}}%

\usepackage{colortbl}
\definecolor{rk000}{HTML}{A8C992}
\definecolor{rk001}{HTML}{A9CA93}
\definecolor{rk002}{HTML}{AACA94}
\definecolor{rk003}{HTML}{ABCB95}
\definecolor{rk004}{HTML}{ADCB95}
\definecolor{rk005}{HTML}{AECC96}
\definecolor{rk006}{HTML}{AFCC97}
\definecolor{rk007}{HTML}{B0CD98}
\definecolor{rk008}{HTML}{B1CD99}
\definecolor{rk009}{HTML}{B2CE9A}
\definecolor{rk010}{HTML}{B3CE9A}
\definecolor{rk011}{HTML}{B5CF9B}
\definecolor{rk012}{HTML}{B6CF9C}
\definecolor{rk013}{HTML}{B7D09D}
\definecolor{rk014}{HTML}{B8D19E}
\definecolor{rk015}{HTML}{B9D19F}
\definecolor{rk016}{HTML}{BAD29F}
\definecolor{rk017}{HTML}{BCD2A1}
\definecolor{rk018}{HTML}{BED4A2}
\definecolor{rk019}{HTML}{C1D5A4}
\definecolor{rk020}{HTML}{C3D6A6}
\definecolor{rk021}{HTML}{C5D7A8}
\definecolor{rk022}{HTML}{C8D8A9}
\definecolor{rk023}{HTML}{CADAAB}
\definecolor{rk024}{HTML}{CDDBAD}
\definecolor{rk025}{HTML}{CFDCAE}
\definecolor{rk026}{HTML}{D1DDB0}
\definecolor{rk027}{HTML}{D4DEB2}
\definecolor{rk028}{HTML}{D6E0B4}
\definecolor{rk029}{HTML}{D9E1B5}
\definecolor{rk030}{HTML}{DBE2B7}
\definecolor{rk031}{HTML}{DDE3B9}
\definecolor{rk032}{HTML}{E0E4BB}
\definecolor{rk033}{HTML}{E2E6BC}
\definecolor{rk034}{HTML}{E4E6BE}
\definecolor{rk035}{HTML}{E6E7BF}
\definecolor{rk036}{HTML}{E7E8C0}
\definecolor{rk037}{HTML}{E9E8C1}
\definecolor{rk038}{HTML}{EAE9C2}
\definecolor{rk039}{HTML}{ECE9C3}
\definecolor{rk040}{HTML}{EDEAC4}
\definecolor{rk041}{HTML}{EFEBC5}
\definecolor{rk042}{HTML}{F1EBC6}
\definecolor{rk043}{HTML}{F2ECC7}
\definecolor{rk044}{HTML}{F4ECC9}
\definecolor{rk045}{HTML}{F5EDCA}
\definecolor{rk046}{HTML}{F7EECB}
\definecolor{rk047}{HTML}{F8EECC}
\definecolor{rk048}{HTML}{FAEFCD}
\definecolor{rk049}{HTML}{FBEFCE}
\definecolor{rk050}{HTML}{FDF0CF}
\definecolor{rk051}{HTML}{FDF0CF}
\definecolor{rk052}{HTML}{FDEFCF}
\definecolor{rk053}{HTML}{FDEFCE}
\definecolor{rk054}{HTML}{FDEECE}
\definecolor{rk055}{HTML}{FDEECE}
\definecolor{rk056}{HTML}{FDEDCE}
\definecolor{rk057}{HTML}{FDEDCE}
\definecolor{rk058}{HTML}{FDECCE}
\definecolor{rk059}{HTML}{FEECCD}
\definecolor{rk060}{HTML}{FEEBCD}
\definecolor{rk061}{HTML}{FEEBCD}
\definecolor{rk062}{HTML}{FEEACD}
\definecolor{rk063}{HTML}{FEEACD}
\definecolor{rk064}{HTML}{FEE9CC}
\definecolor{rk065}{HTML}{FEE9CC}
\definecolor{rk066}{HTML}{FEE8CC}
\definecolor{rk067}{HTML}{FEE8CC}
\definecolor{rk068}{HTML}{FEE7CC}
\definecolor{rk069}{HTML}{FEE6CB}
\definecolor{rk070}{HTML}{FEE6CB}
\definecolor{rk071}{HTML}{FDE5CA}
\definecolor{rk072}{HTML}{FDE4CA}
\definecolor{rk073}{HTML}{FDE3CA}
\definecolor{rk074}{HTML}{FDE3C9}
\definecolor{rk075}{HTML}{FDE2C9}
\definecolor{rk076}{HTML}{FDE1C9}
\definecolor{rk077}{HTML}{FDE1C8}
\definecolor{rk078}{HTML}{FDE0C8}
\definecolor{rk079}{HTML}{FDDFC8}
\definecolor{rk080}{HTML}{FCDEC7}
\definecolor{rk081}{HTML}{FCDEC7}
\definecolor{rk082}{HTML}{FCDDC6}
\definecolor{rk083}{HTML}{FCDCC6}
\definecolor{rk084}{HTML}{FBDBC5}
\definecolor{rk085}{HTML}{FBD9C4}
\definecolor{rk086}{HTML}{FAD8C3}
\definecolor{rk087}{HTML}{F9D6C2}
\definecolor{rk088}{HTML}{F8D4C1}
\definecolor{rk089}{HTML}{F7D2C0}
\definecolor{rk090}{HTML}{F6D1BF}
\definecolor{rk091}{HTML}{F6CFBE}
\definecolor{rk092}{HTML}{F5CDBD}
\definecolor{rk093}{HTML}{F4CCBC}
\definecolor{rk094}{HTML}{F3CABB}
\definecolor{rk095}{HTML}{F2C8BA}
\definecolor{rk096}{HTML}{F1C7B9}
\definecolor{rk097}{HTML}{F1C5B8}
\definecolor{rk098}{HTML}{F0C3B7}
\definecolor{rk099}{HTML}{EFC2B6}
\definecolor{rk100}{HTML}{EEC0B5}
\newsavebox{\isotabbox}\newlength{\isotabwidth}
\definecolor{oursrow}{gray}{0.93}
\newcommand{\best}[1]{\textbf{#1}}

\newcommand{\fail}{\textcolor{gray!70}{fail}}
\newcommand{\metric}[1]{{\scriptsize\textcolor{black!70}{#1}}}

\definecolor{cvprblue}{rgb}{0.21,0.49,0.74}
\usepackage[pagebackref,breaklinks,colorlinks,allcolors=cvprblue]{hyperref}

\def\confName{CVPR}
\def\confYear{2027}

\title{FoundationCov: Making Geometric Foundation Models Probabilistic with Predictive Covariances}
\author{%
  Ruogu Li\textsuperscript{1,*}\quad
  Xiang Fei\textsuperscript{1,*,\textdagger}\quad
  Jiting Cai\textsuperscript{1,*,\textdagger}\quad
  Hyeokjoon Kwon\textsuperscript{1}\\[2pt]
  Yutian Chen\textsuperscript{2}\quad
  Yuchen Zhang\textsuperscript{1}\quad
  Sebastian Scherer\textsuperscript{1}\quad
  Wenshan Wang\textsuperscript{1}\\[6pt]
  {\normalfont\normalsize%
    \textsuperscript{1}Carnegie Mellon University\qquad
    \textsuperscript{2}Stanford University}%
}
\hypersetup{pdfauthor={Ruogu Li, Xiang Fei, Jiting Cai, Hyeokjoon Kwon, Yutian Chen, Yuchen Zhang, Sebastian Scherer, Wenshan Wang},
  pdftitle={FoundationCov: Making Geometric Foundation Models Probabilistic with Predictive Covariances}}

\begin{document}
\twocolumn[{%
    \renewcommand\twocolumn[1][]{#1}%
    \maketitle
    \vspace{-10pt}

    \begin{center}
        \includegraphics[width=\textwidth]{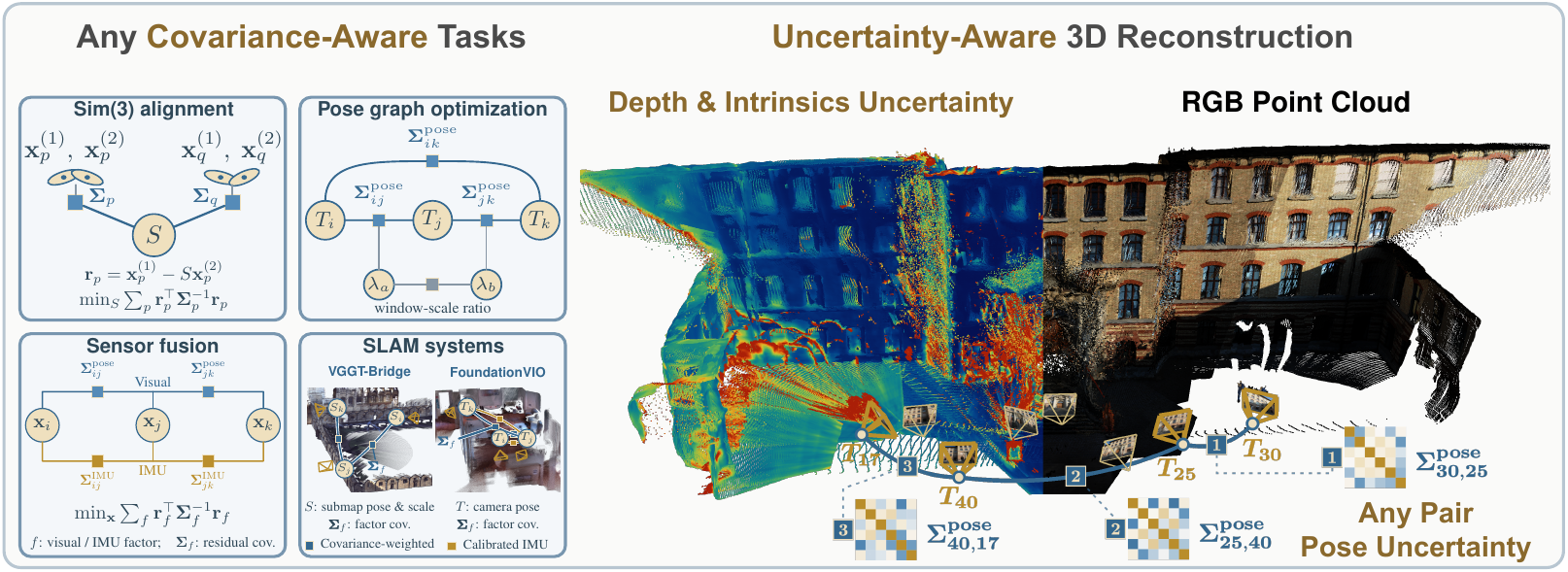}
        \vspace{-6pt}
        \captionof{figure}{\textbf{FoundationCov equips geometric foundation models with covariance for downstream estimation.}
        Our framework predicts depth, relative pose, and intrinsics covariance while preserving the original geometric predictions.
        \textbf{Left:} The predicted covariances support probabilistic weighting and uncertainty propagation in multiple downstream tasks.
        \textbf{Right:} A reconstructed scene visualized with point uncertainty and RGB colors, together with full relative pose covariances for selected frames.}
    \label{fig:teaser}
    \end{center}
    \vspace{6pt}
}]

\begingroup
\renewcommand{\thefootnote}{*}
\footnotetext[1]{Equal contribution.\quad\textsuperscript{\textdagger}Project leads.}
\endgroup
\begin{abstract}
Feed-forward geometric foundation models demonstrate remarkable capabilities in recovering 3D scene geometry and camera poses from a sequence of images. However, despite their impressive accuracy, these models typically provide deterministic predictions with limited characterization of their uncertainty, making it difficult to reliably incorporate their outputs into downstream geometric estimation and sensor fusion pipelines. In this work, we introduce \textbf{FoundationCov}, a general covariance prediction framework for feed-forward geometric foundation models that jointly estimates dense pixelwise depth uncertainty, full 6-DoF relative pose covariance, and camera intrinsics covariance.
Extensive experiments demonstrate that FoundationCov produces covariances that reflect the observed errors on unseen datasets without test-time calibration and improves downstream performance in uncertainty-aware multi-view reconstruction and visual-inertial estimation. Our work takes a step toward transforming feed-forward geometric foundation models from standalone reconstruction systems into probabilistic measurement providers for downstream 3D perception and estimation. Our code and model checkpoints will be publicly available.
\end{abstract}

\section{Introduction}

Feed-forward geometric foundation models (GFMs) have changed how 3D structure is recovered from images.
Building on the direct regression paradigm introduced by DUSt3R~\cite{wang2024dust3r}, models such as VGGT~\cite{wang2025vggt}, VGGT-$\Omega$~\cite{wang2026vggtomega} and Depth Anything 3~\cite{lin2025depthanything3} predict camera poses, intrinsics, and dense geometry from a set of images in a single forward pass.
These models combine strong geometric prediction capabilities with efficient inference, achieving impressive results in camera pose estimation, depth estimation, and multi-view reconstruction.
This combination of accuracy and efficiency has led to their growing adoption in 3D perception, mapping, and state estimation systems~\cite{deng2026vggt,choi2026vggtbridge,rhee2026foundationvio}.
Within these systems, the predictions of geometric feed-forward models are rarely used in isolation.
They are fused with measurements from other sensors, predictions from different views or time steps, and prior constraints on scene structure and motion.
Such fusion is often formulated as maximum a posteriori (MAP) estimation, in which the influence of each constraint is determined by its associated error distribution.
Effective fusion therefore requires both geometric predictions and uncertainty estimates that characterize the distributions of their prediction errors.

However, advances in geometric prediction have not been accompanied by probabilistic uncertainty modeling.
Camera parameters, including poses and intrinsics, are typically predicted without corresponding uncertainty.
The confidence maps that accompany depth predictions are effective for discarding unreliable regions~\cite{hillemann2026uncertainty,wang2026vggtomega}, but they do not define an error distribution.
Consequently, the predictions come without a measurement likelihood, which is what a MAP estimator needs to weight each constraint against the others.
In geometric estimation, this likelihood is conventionally specified by a covariance, which characterizes the magnitude and directional structure of a measurement’s error. Lacking such a covariance, downstream systems weight these predictions with constant noise models or heuristics such as confidence scores and point counts~\cite{deng2026vggt,choi2026vggtbridge,rhee2026foundationvio}, which amounts to MAP estimation under an assumed error model that is not calibrated to the actual errors.

In this paper, we replace these assumed error models with learned covariance.
We present \textbf{\method{}}, which turns the outputs of a pretrained GFM into probabilistic measurements by attaching to each prediction a learned covariance, defined in the form a downstream estimator consumes.
Specifically, it predicts per-pixel depth variance, a full $6\times6$ covariance for the relative pose of any pair of frames, and covariance for camera intrinsics, which
could be seamlessly integrated to the established probabilistic geometric estimation frameworks.
We observe that the intermediate features of a GFM already carry information about the errors of its predictions. FoundationCov therefore keeps the GFM frozen and adds only lightweight prediction heads on these features, which leaves the geometric predictions themselves unchanged. The same architecture applies to multiple GFMs, such as VGGT, VGGT-$\Omega$, and Depth Anything 3.

We validate \method{} at three levels of increasing complexity.
First, on unseen datasets, the predicted covariances reflect the distribution of the observed errors.
Second, in isolated Sim(3) alignment, loop closure, and visual--inertial initialization, they improve accuracy over equal and heuristic weights.
Third, integrated into complete visual and visual--inertial SLAM systems~\cite{rhee2026foundationvio,choi2026vggtbridge}, they reduce trajectory error relative to the default weighting by 45\% on average over four diverse datasets without any retuning, \eg, by 37\% on the standard EuRoC benchmark and by 68\% on the challenging, large-scale Oxford Spires dataset.

In summary, our main contributions are:
\begin{itemize}
  \setlength{\itemsep}{2pt}
  \setlength{\parskip}{0pt}
  \item We turn the outputs of GFMs into probabilistic measurements by defining and learning error distributions for relative pose, depth, and focal length in the form downstream estimators consume, including a full $6\times6$ covariance for the relative pose of any frame pair.
  \item We show that the frozen representations of pretrained geometric foundation models already encode their own error distributions. We introduce \method{}, a lightweight architecture that predicts covariances for VGGT, VGGT-$\Omega$, and Depth Anything 3 without altering any prediction.
  \item We show that the predicted covariances reflect the observed errors on unseen datasets and that, used without retuning in place of equal or heuristic weights, they improve three isolated estimation tasks and two complete visual and visual--inertial SLAM systems, reducing the trajectory error by 45\% on average over four datasets.
\end{itemize}

\section{Related Work}
\label{sec:related}

\paragraph{Feed-forward geometric foundation models.}
Given an uncalibrated image pair, DUSt3R~\cite{wang2024dust3r} regresses dense 3D geometry directly, and MASt3R~\cite{leroy2024mast3r} extends it with dense matching.
Multi-view successors such as VGGT~\cite{wang2025vggt}, $\pi^3$~\cite{wang2025pi3}, MapAnything~\cite{keetha2026mapanything}, Depth Anything 3~\cite{lin2025depthanything3}, and VGGT-$\Omega$~\cite{wang2026vggtomega} predict cameras and dense geometry for many frames in a single forward pass.
Because these models operate on limited image windows and rely on visual input alone, larger systems treat them as front-ends for optimization: they align local reconstructions and improve global consistency through loop closure~\cite{murai2025mast3rslam,maggio2026vggt,deng2026vggt,choi2026vggtbridge}, or fuse predicted camera poses with inertial measurements~\cite{rhee2026foundationvio}.
These optimization objectives require appropriate weighting of residuals derived from the model predictions. For example, VGGT-Bridge and FoundationVIO set these weights from confidence scores, point-count heuristics, or constant noise models.
\method{} is complementary to this line of work: it freezes the GFMs and provides learned error models for downstream optimization.

\paragraph{Uncertainty in feed-forward reconstruction.}
DUSt3R trains a per-pixel confidence jointly with its regression loss, in the manner of learned loss attenuation~\cite{kendall2017uncertainties}, and its successors inherit this output.
These confidences are effective for filtering unreliable regions~\cite{hillemann2026uncertainty,wang2026vggtomega,chen2025come}, but they are scores without units or orientation rather than variances of a stated error.
Moreover, these models do not provide uncertainty estimates for their predicted camera poses and intrinsics.
Hillemann~\etal~\cite{hillemann2026uncertainty} show that VGGT's confidence maps identify inaccurate points only weakly, and that better uncertainty estimates could substantially improve its reconstructions.
\method{} instead predicts covariances for the relative pose of any frame pair, for depth, and for focal length. It applies to multiple GFMs, and its covariances can be used directly in downstream estimators.

\paragraph{Covariance for geometric estimation.}
Probabilistic estimators weight each measurement by its inverse covariance, which encodes how much the measurement should be trusted.
Classical systems rely on predefined error models: VINS-Mono~\cite{qin2018vins} and OKVIS~\cite{leutenegger2015okvis} weight all visual residuals by a fixed pixel noise, ORB-SLAM3~\cite{campos2021orbslam3} sets it by the image-pyramid level at which each feature is detected, and inertial residuals are weighted by covariances propagated from calibrated IMU noise parameters.
Such rules cannot accurately reflect how reliable each measurement is in its local context, such as texture and scene geometry.
Learning-based methods instead learn the covariance of each measurement directly from data.
D3VO~\cite{yang2020d3vo} learns photometric uncertainty to weight the residuals of direct visual odometry. MAC-VO~\cite{qiu2025macvo} learns covariances of matched 3D points for pose optimization, RoMa~v2~\cite{edstedt2026romav2} predicts the covariance of dense matches for geometric refinement, and MAC-I$^2$~\cite{fei2026maci2} uses learned covariances for visual--inertial fusion.
These methods are designed for the measurements of a specific front-end, such as photometric residuals, feature matches, or IMU integration, and do not apply to GFMs, which predict pose, intrinsics, and depth from images in a single forward pass.
We instead learn covariances for these GFM predictions, so that downstream estimators can use them directly.

\section{Method}
\label{sec:method}
\begin{figure*}[t]
    \centering
    \includegraphics[width=0.95\textwidth]{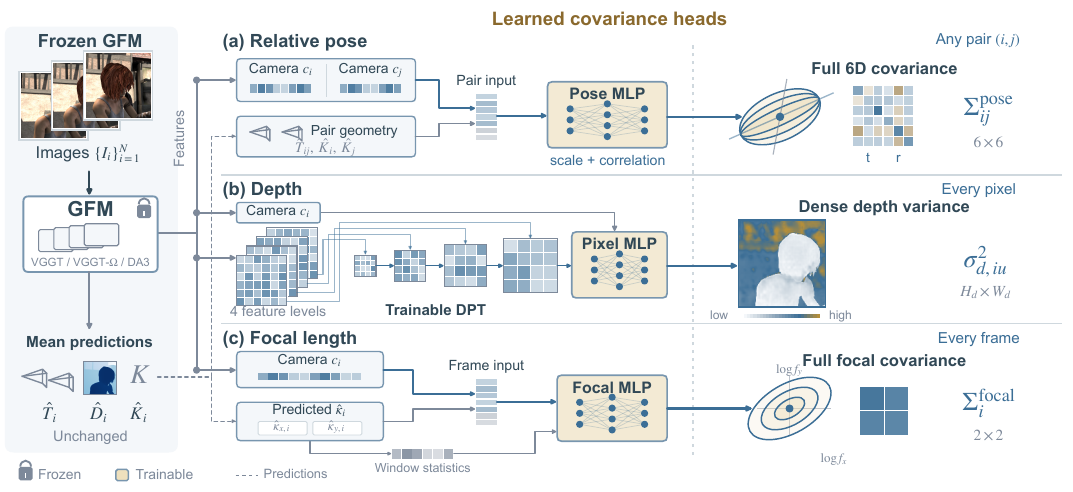}
    \caption{\textbf{FoundationCov architecture.}
    Frozen camera features condition the pose and focal covariance heads; a separate trainable DPT decodes patch features for pixelwise depth variance.}
    \label{fig:architecture}
\end{figure*}

\method{} turns the deterministic predictions of a pretrained feed-forward geometric foundation model (GFM) into probabilistic measurements. We keep the model frozen and train lightweight heads that predict their covariances.

\subsection{From Images to Geometric Predictions}
\label{sec:method_geometry}

A GFM maps a window of images $\mathcal I=\{I_i\in\mathbb R^{H\times W\times3}\}_{i=1}^{N}$ to
\begin{equation}
 \mathcal G_\theta(\mathcal I)
 =\{\hat T_i,\hat D_i,\hat K_i\}_{i=1}^{N},
 \label{eq:geometric_model}
\end{equation}
where $\hat T_i\in\mathrm{SE}(3)$ is the camera-to-world pose of frame $i$ in a common reconstruction frame, $\hat D_i$ its depth map, and $\hat K_i$ its pinhole intrinsics.
From monocular images, translations and depths are recovered only up to a global scale per window, which we call the reconstruction scale.

\subsection{Probabilistic Geometric Predictions}
\label{sec:method_errors}

We define error distributions for three outputs of a GFM, so that downstream estimators can weight and fuse them: the relative camera pose between two arbitrary frames, the per-pixel depth, and the focal length.

\paragraph{Distributions of predictions.}
We describe each prediction by the distribution of its error $e\in\mathbb R^k$ with respect to the ground truth; its definition for relative pose, depth, and focal length is given in the corresponding paragraphs of this section.
We model this error as a zero-mean Gaussian,
\begin{equation}
 p_\psi(e\mid\mathcal I)
 =\mathcal N\bigl(e;0,\Sigma_\psi(\mathcal I)\bigr),
 \label{eq:error_distribution}
\end{equation}
whose covariance $\Sigma_\psi(\mathcal I)\in\mathbb R^{k\times k}$ is predicted from the frozen features by a head parameterized by $\psi$. A downstream estimator uses $\Sigma_\psi^{-1}$ to weight the corresponding residual, in place of constant noise or confidence scores.

\paragraph{Reconstruction scale.}
The GFM's translations and depths are in its reconstruction scale, which generally differs from the ground truth scale. To compare the two, we rescale the ground truth rather than the prediction. The covariance then describes the prediction in its own units.
For each training window, we fit a single scale $s>0$ between the predicted and ground-truth camera positions and divide the ground-truth translations and depths by $s$. A superscript $\star$ denotes ground truth after this rescaling.

\paragraph{Relative pose.}
We model the error of the relative pose between any frames $i$ and $j$ of the window, rather than for a fixed choice of pairs. Poses relative to the first camera and poses between consecutive frames are therefore special cases, and the downstream estimator can use whichever pairs it needs.
With $\hat T_{ij}=\hat T_i^{-1}\hat T_j$ mapping points from camera $j$ into camera $i$, the error is
\begin{equation}
 \xi_{ij}=\operatorname{Log}\bigl(\hat T_{ij}^{-1}T^\star_{ij}\bigr)
 \in\mathbb R^6.
 \label{eq:pose_error}
\end{equation}
The pose head predicts its full covariance $\Sigma^{\mathrm{pose}}_{ij}\in\mathbb R^{6\times6}$, including the coupling between translation and rotation.

\paragraph{Depth.}
At each pixel $u$ of frame $i$, we model the additive depth error $e^d_{iu}=d_i^\star(u)-\hat d_i(u)$ as a zero-mean Gaussian with variance $\sigma^2_{d,iu}$.
The head outputs a log-variance parameter $\ell^d_{iu}=\log v_{iu}$, from which we obtain
\begin{equation}
 \sigma^2_{d,iu}=\hat d_i(u)^2\exp(\ell^d_{iu}).
 \label{eq:depth_variance}
\end{equation}
Here $v_{iu}$ is the variance of the normalized error $e^d_{iu}/\hat d_i(u)$; $\sigma^2_{d,iu}$ is the depth variance in squared reconstruction units used by downstream estimators.

\paragraph{Focal length.}
We represent the intrinsics of frame $i$ by its log focal lengths normalized by the image size, $\kappa_i=(\log(f_{x,i}/W),\log(f_{y,i}/H))^\top$, so that the error $e^K_i=\hat\kappa_i-\kappa_i^\star$ is a relative focal error independent of the image resolution.
The focal head predicts its covariance $\Sigma_i^{\mathrm{focal}}\in\mathbb R^{2\times2}$.

\subsection{Covariance Heads on Frozen Features}
\label{sec:method_features}

The three heads read frozen features (\cref{fig:architecture}).

\paragraph{Pose head.}
For a frame pair $(i,j)$, an MLP $g_{\mathrm{pose}}$ with two residual blocks predicts the parameters of the covariance,
\begin{equation}
 (\ell_{ij},C_{ij})=g_{\mathrm{pose}}\bigl(c_i,c_j,\gamma_{ij}\bigr),
 \label{eq:pose_head}
\end{equation}
where $c_i,c_j$ project the final geometric-transformer camera tokens and mean register tokens, when available.
The geometry input $\gamma_{ij}$ contains $\hat T_{ij}$, both intrinsics, and the translation and rotation magnitudes.

The outputs of $g_{\mathrm{pose}}$ are six log variances $\ell_{ij}\in\mathbb R^6$ and a correlation matrix $C_{ij}$, which define
\begin{equation}
 \Sigma^{\mathrm{pose}}_{ij}=S_{ij}C_{ij}S_{ij},\qquad
 S_{ij}=\operatorname{diag}\bigl(\exp(\ell_{ij}/2)\bigr).
 \label{eq:pose_covariance}
\end{equation}
We define $C_{ij}$ from fifteen unconstrained parameters such that it is always a valid correlation matrix (\cref{eq:supp_pose_parameterization}). $\Sigma^{\mathrm{pose}}_{ij}$ is therefore positive definite, and $\exp(\ell_{ij})$ are its marginal variances.

\paragraph{Depth head.}
We obtain a feature map $F_i$ from a trainable copy of the feature-processing layers of the GFM's depth decoder, a dense prediction transformer (DPT)~\cite{ranftl2021dpt}, applied to the same four levels of patch features; the original decoder, and hence $\hat D_i$, stays frozen.
A pointwise MLP $g_{\mathrm{depth}}$ then predicts the log variance of \cref{eq:depth_variance} at each pixel,
\begin{equation}
 \ell^d_{iu}=g_{\mathrm{depth}}\bigl(F_i(u),Pc_i^{\mathrm{head}}\bigr),
 \label{eq:depth_head}
\end{equation}
where $Pc_i^{\mathrm{head}}$ projects the final camera-head features to provide frame-level context.

\paragraph{Focal head.}
For each frame $i$, an MLP $g_{\mathrm{focal}}$ predicts a lower-triangular factor of the covariance,
\begin{equation}
 \begin{aligned}
 L_i&=g_{\mathrm{focal}}\bigl(c_i,\hat\kappa_i,a_i\bigr),\\
 \Sigma^{\mathrm{focal}}_i&=L_iL_i^\top+\epsilon_K^2I,
 \end{aligned}
 \label{eq:focal_head}
\end{equation}
from the frame summary $c_i$, the predicted log focal lengths $\hat\kappa_i$, and a few statistics $a_i$ of the focal lengths predicted for the other frames of the window (\cref{sec:supp_method}); $\epsilon_K^2$ is a small variance floor.

\subsection{Training}
\label{sec:method_training}

We train the heads of each GFM on its frozen features by minimizing the Gaussian negative log-likelihood of the errors of \cref{sec:method_errors},
\begin{equation}
 \ell(e,\Sigma)=\tfrac12\bigl(e^\top\Sigma^{-1}e+
 \log\det\Sigma+k\log(2\pi)\bigr),
 \label{eq:gaussian_nll}
\end{equation}
where $k$ is the dimension of $e$.
No ground-truth covariance is required, only the error $e$: in expectation, this loss is minimized when $\Sigma$ equals the second moment of the error given the features, so it rewards calibration and not merely a correct ranking.
For the pose head, each window contributes its adjacent frame pairs and a sample of non-adjacent ones.

\section{Experiments}
\label{sec:exp}

We first test whether the predicted covariances match the observed geometric errors on held-out data (\cref{sec:exp_calibration}).
We then evaluate their utility in three estimation problems in isolation, where only the weighting changes (\cref{sec:exp_isolated}), and in two complete systems, where the covariances replace the released weights without any retuning (\cref{sec:exp_systems}).

\subsection{Experimental Setup}
\label{sec:exp_setup}

\paragraph{GFMs.}
We equip VGGT~\cite{wang2025vggt}, VGGT-$\Omega$~\cite{wang2026vggtomega} and Depth Anything 3 (DA3, Large)~\cite{lin2025depthanything3} with \method{}.

\paragraph{Training details.}
Each model is trained for 17,000 steps on eight NVIDIA H200 GPUs, with training windows of $N\in\{10,16,24,32\}$ frames (optimizer and schedule in \cref{sec:supp_method}).

\paragraph{Training data.}
We train the covariance heads separately for all three backbones on the same 16
datasets: Aria Digital Twin (ADT)~\cite{pan2023aria}, Argoverse~2~\cite{wilson2023argoverse}, BlendedMVS~\cite{yao2020blendedmvs}, DL3DV~\cite{ling2024dl3dv},
DynamicReplica~\cite{karaev2023dynamicstereo}, Hypersim~\cite{roberts2021hypersim}, Mapillary Planet-Scale Depth (MPSD)~\cite{antequera2020mapillary}, Mid-Air~\cite{fonder2019mid},
MVS-Synth~\cite{huang2018deepmvs}, ParallelDomain4D~\cite{van2024generative}, SAIL-VOS~3D~\cite{hu2021sail}, ScanNet~\cite{dai2017scannet}, ScanNet++~\cite{yeshwanth2023scannet++}, TartanAir~V2~\cite{wang2020tartanair},
UnrealStereo4K~\cite{tosi2021smd}, and Virtual KITTI~2~\cite{cabon2020virtual}.

\paragraph{Evaluation data.}
For covariance calibration, we use TUM-RGBD~\cite{sturm2012tum}
and ETH3D~\cite{schoeps2017eth3d} for relative pose;
Spring~\cite{mehl2023spring} and 7-Scenes~\cite{shotton2013scenes} for depth;
and nuScenes~\cite{caesar2020nuscenes} and
ScaRF~\cite{zhang2026scarfslam} for focal length.
For downstream evaluation, we use four zero shot visual-inertial datasets: EuRoC MAV~\cite{burri2016euroc},
Aria~\cite{engel2023projectaria},
ScaRF~\cite{zhang2026scarfslam} %
and Oxford Spires~\cite{tao2026oxford}.

\paragraph{Metrics.}
We assess the predicted covariances with the average normalized estimation error squared (ANEES)~\cite{bar2001estimation} and the area under the calibration error curve (AUCE)~\cite{gustafsson2020evaluating}.
For the $n$ errors $e_i\in\mathbb R^k$ of a window, with predicted covariances $\Sigma_i$,
\begin{equation}
  \mathrm{ANEES}=\frac{1}{nk}\sum_{i=1}^{n}e_i^\top\Sigma_i^{-1}e_i,\quad
  \mathrm{AUCE}=\int_0^1\!\big|\hat p(p)-p\big|\,\mathrm{d}p.
  \label{eq:exp_calibration_metrics}
\end{equation}
ANEES is one when the covariances agree with the errors in scale; values above one indicate overconfident and values below one conservative covariances.
We report its median over windows.
In AUCE, $\hat p(p)$ is the empirical coverage: the fraction of errors inside the confidence region to which the predicted Gaussian assigns probability $p$.
AUCE is zero when the empirical coverage equals the nominal coverage $p$ everywhere; we compute it from the mean coverage curve of each dataset.
For a relative Sim(3) $\hat S=(\hat s,\hat{\mathbf{R}},\hat{\mathbf{t}})$ with ground truth $S=(s,\mathbf{R},\mathbf{t})$ we report
\begin{equation}
  \begin{aligned}
    \mathrm{RRE} &= \angle\big(\hat{\mathbf{R}}\mathbf{R}^{\top}\big),\quad
    \mathrm{RTE} = \|\hat{\mathbf{t}}-\mathbf{t}\|,\\
    \mathrm{RSE} &= 100\,\big|\log(\hat s/s)\big|.
  \end{aligned}
  \label{eq:exp_rel}
\end{equation}
in degrees, centimeters (in ground-truth scale) and percent.
For trajectories we report the ATE (m) after Umeyama alignment, Sim(3) for vision-only and SE(3) for visual-inertial estimates, and the translational and rotational RPE over 1\,s intervals~\cite{sturm2012tum} (m/s, $^\circ$/s).
For these downstream metrics, dataset values are means over the sequences, and for the relative Sim(3) errors over all edges.

\subsection{Calibration of Predicted Covariances}
\label{sec:exp_calibration}

\begin{table}[t]
\centering
\scriptsize
\setlength{\tabcolsep}{1.2pt}
\caption{\textbf{Zero-shot calibration.} Median window ANEES (ideal: 1) and AUCE (ideal: 0).}
\label{tab:calibration}
\begin{tabular}{@{}lrrrrrr@{}}
\toprule
& \multicolumn{2}{c}{VGGT-$\Omega$} & \multicolumn{2}{c}{VGGT} & \multicolumn{2}{c}{DA3} \\
\cmidrule(lr){2-3}\cmidrule(lr){4-5}\cmidrule(lr){6-7}\addlinespace[0.3ex]
Dataset & ANEES & AUCE$\downarrow$ & ANEES & AUCE$\downarrow$ & ANEES & AUCE$\downarrow$ \\
\midrule
\multicolumn{7}{@{}l}{\textit{Relative pose}} \\
TUM-RGBD~\cite{sturm2012tum} & 0.817 & 0.061 & 0.860 & 0.111 & 1.058 & 0.049 \\
ETH3D~\cite{schoeps2017eth3d} & 1.126 & 0.049 & 0.706 & 0.139 & 0.730 & 0.140 \\
\midrule
\multicolumn{7}{@{}l}{\textit{Depth}} \\
Spring~\cite{mehl2023spring} & 0.540 & 0.039 & 2.062 & 0.114 & 1.155 & 0.133 \\
7-Scenes~\cite{shotton2013scenes} & 0.538 & 0.144 & 1.150 & 0.049 & 2.442 & 0.059 \\
\midrule
\multicolumn{7}{@{}l}{\textit{Focal length}} \\
nuScenes~\cite{caesar2020nuscenes} & 0.573 & 0.036 & 0.614 & 0.046 & 1.178 & 0.130 \\
ScaRF~\cite{zhang2026scarfslam} & 1.098 & 0.082 & 1.752 & 0.163 & 1.149 & 0.103 \\
\bottomrule
\end{tabular}
\end{table}

\begin{figure*}[t]
  \centering
  \includegraphics[width=0.75\textwidth]{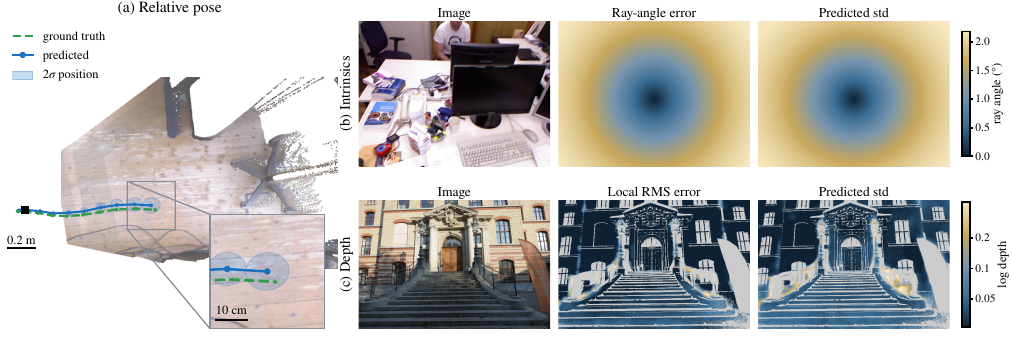}
  \caption{
  \textbf{Predicted covariances on unseen data (VGGT-$\Omega$).}
(a) Ground-truth and predicted camera centers with predicted $2\sigma$
position ellipses on TUM RGB-D; each ellipse shows the covariance of
frame $i$ relative to frame 0, $\Sigma^{\mathrm{pose}}_{0i}$.
(b) Ray-angle errors due to focal-length estimation and the corresponding
predicted standard deviations on TUM RGB-D.
(c)  Local RMS log-depth errors and predicted standard deviations on
ETH3D; calibrated standard deviations describe RMS errors rather than
individual residuals~\cite{levi2022evaluating}.
Gray indicates missing ground truth.
  }
  \label{fig:covariances}
\end{figure*}

\paragraph{Results.}
On datasets unseen in training and without any test-time calibration, median ANEES stays between 0.54 and 2.44 across the three GFMs (\cref{tab:calibration}), \ie, the predicted standard deviations are within a factor of about 1.6 of the observed errors.
AUCE lies between 0.04 and 0.16, with a mean of 0.09 that is nearly identical for pose, depth, and focal length, \ie, the empirical coverage of the predicted confidence regions deviates from the nominal coverage by about nine percentage points on average.
Together, the two metrics indicate that the predicted covariances reflect both the scale and the distribution of the errors on unseen data, which is what allows them to be used directly as weights in the downstream experiments below.
Fig.~\ref{fig:covariances} illustrates this on single examples: the ground-truth camera positions stay within the predicted $2\sigma$ ellipses (a), and the predicted standard deviations follow the magnitude and spatial pattern of the ray-angle errors (b) and of the local depth errors (c).

\begin{table*}[t]
\centering
\footnotesize
\setlength{\tabcolsep}{3.4pt}
\caption{
\textbf{Isolated estimation tasks.}
We evaluate \method{} on (a) Sim(3) alignment, (b) loop closure,
and (c) visual--inertial initialization using VGGT-$\Omega$, VGGT,
and DA3.
Best per backbone in \textbf{bold}.}
\sbox{\isotabbox}{\begin{tabular}{@{}llcccccccccccc@{}}
\toprule
 & & \multicolumn{3}{c}{EuRoC} & \multicolumn{3}{c}{Aria} & \multicolumn{3}{c}{ScaRF} & \multicolumn{3}{c}{Spires} \\
\midrule
\multicolumn{2}{@{}l}{\textit{(a) Sim(3) alignment of overlapping submaps}} & \metric{RRE} & \metric{RTE} & \metric{RSE} & \metric{RRE} & \metric{RTE} & \metric{RSE} & \metric{RRE} & \metric{RTE} & \metric{RSE} & \metric{RRE} & \metric{RTE} & \metric{RSE} \\
\cmidrule(r){1-2}\cmidrule(lr){3-5}\cmidrule(lr){6-8}\cmidrule(lr){9-11}\cmidrule(lr){12-14}
 & Identity & \cellcolor{rk100}0.95 & \cellcolor{rk100}12.2 & \cellcolor{rk100}9.27 & \cellcolor{rk100}0.94 & \cellcolor{rk100}6.2 & \cellcolor{rk100}2.80 & \cellcolor{rk100}1.02 & \cellcolor{rk100}10.7 & \cellcolor{rk100}3.87 & \cellcolor{rk100}0.51 & \cellcolor{rk100}22.1 & \cellcolor{rk100}4.33 \\
VGGT-$\Omega$ & Heuristic & \cellcolor{rk050}0.86 & \cellcolor{rk050}9.5 & \cellcolor{rk050}8.77 & \cellcolor{rk050}0.88 & \cellcolor{rk050}4.9 & \cellcolor{rk050}2.13 & \cellcolor{rk050}0.99 & \cellcolor{rk050}9.6 & \cellcolor{rk050}3.41 & \cellcolor{rk050}0.45 & \cellcolor{rk050}12.4 & \cellcolor{rk050}3.09 \\
\rowcolor{oursrow} & \method{} & \cellcolor{rk000}\best{0.61} & \cellcolor{rk000}\best{6.6} & \cellcolor{rk000}\best{6.18} & \cellcolor{rk000}\best{0.79} & \cellcolor{rk000}\best{4.2} & \cellcolor{rk000}\best{1.96} & \cellcolor{rk000}\best{0.85} & \cellcolor{rk000}\best{8.2} & \cellcolor{rk000}\best{3.31} & \cellcolor{rk000}\best{0.32} & \cellcolor{rk000}\best{4.8} & \cellcolor{rk000}\best{1.80} \\
\cmidrule(l){2-14}
 & Identity & \cellcolor{rk100}2.62 & \cellcolor{rk100}33.6 & \cellcolor{rk100}16.7 & \cellcolor{rk100}9.35 & \cellcolor{rk100}38.1 & \cellcolor{rk100}19.6 & \cellcolor{rk100}9.66 & \cellcolor{rk100}38.0 & \cellcolor{rk100}15.5 & \cellcolor{rk100}1.91 & \cellcolor{rk100}42.2 & \cellcolor{rk100}8.74 \\
VGGT & Heuristic & \cellcolor{rk050}2.56 & \cellcolor{rk050}30.9 & \cellcolor{rk050}16.3 & \cellcolor{rk050}9.32 & \cellcolor{rk050}37.9 & \cellcolor{rk050}19.5 & \cellcolor{rk050}9.62 & \cellcolor{rk050}37.2 & \cellcolor{rk050}15.4 & \cellcolor{rk050}1.89 & \cellcolor{rk050}40.1 & \cellcolor{rk050}8.36 \\
\rowcolor{oursrow} & \method{} & \cellcolor{rk000}\best{1.62} & \cellcolor{rk000}\best{16.2} & \cellcolor{rk000}\best{11.7} & \cellcolor{rk000}\best{7.84} & \cellcolor{rk000}\best{33.6} & \cellcolor{rk000}\best{18.7} & \cellcolor{rk000}\best{7.10} & \cellcolor{rk000}\best{28.5} & \cellcolor{rk000}\best{13.6} & \cellcolor{rk000}\best{1.33} & \cellcolor{rk000}\best{22.3} & \cellcolor{rk000}\best{5.85} \\
\cmidrule(l){2-14}
 & Identity & \cellcolor{rk100}1.40 & \cellcolor{rk100}16.2 & \cellcolor{rk100}12.2 & \cellcolor{rk100}2.63 & \cellcolor{rk100}11.4 & \cellcolor{rk100}3.71 & \cellcolor{rk100}3.58 & \cellcolor{rk100}17.6 & \cellcolor{rk100}6.34 & \cellcolor{rk100}0.99 & \cellcolor{rk100}20.4 & \cellcolor{rk100}3.94 \\
DA3 & Heuristic & \cellcolor{rk050}1.26 & \cellcolor{rk050}12.5 & \cellcolor{rk050}11.6 & \cellcolor{rk050}2.53 & \cellcolor{rk050}10.2 & \cellcolor{rk000}\best{3.16} & \cellcolor{rk050}3.55 & \cellcolor{rk050}17.0 & \cellcolor{rk050}6.30 & \cellcolor{rk050}0.93 & \cellcolor{rk050}14.1 & \cellcolor{rk050}2.97 \\
\rowcolor{oursrow} & \method{} & \cellcolor{rk000}\best{0.86} & \cellcolor{rk000}\best{8.0} & \cellcolor{rk000}\best{7.97} & \cellcolor{rk000}\best{2.16} & \cellcolor{rk000}\best{8.9} & \cellcolor{rk050}3.17 & \cellcolor{rk000}\best{2.55} & \cellcolor{rk000}\best{14.8} & \cellcolor{rk000}\best{5.54} & \cellcolor{rk000}\best{0.72} & \cellcolor{rk000}\best{9.2} & \cellcolor{rk000}\best{2.06} \\
\midrule
\multicolumn{2}{@{}l}{\textit{(b) Loop closure}} & \metric{ATE} & \metric{RPE$_t$} & \metric{RPE$_r$} & \metric{ATE} & \metric{RPE$_t$} & \metric{RPE$_r$} & \metric{ATE} & \metric{RPE$_t$} & \metric{RPE$_r$} & \metric{ATE} & \metric{RPE$_t$} & \metric{RPE$_r$} \\
\cmidrule(r){1-2}\cmidrule(lr){3-5}\cmidrule(lr){6-8}\cmidrule(lr){9-11}\cmidrule(lr){12-14}\addlinespace[0.3ex]
 & \textcolor{black!70}{\textit{no loop closure}} & \textcolor{black!70}{0.348} & \textcolor{black!70}{0.082} & \textcolor{black!70}{0.375} & \textcolor{black!70}{0.928} & \textcolor{black!70}{0.094} & \textcolor{black!70}{0.488} & \textcolor{black!70}{1.021} & \textcolor{black!70}{0.167} & \textcolor{black!70}{0.701} & \textcolor{black!70}{11.16} & \textcolor{black!70}{0.432} & \textcolor{black!70}{0.214} \\
 & Identity & \cellcolor{rk050}0.315 & \cellcolor{rk050}0.150 & \cellcolor{rk100}0.395 & \cellcolor{rk100}2.059 & \cellcolor{rk050}0.153 & \cellcolor{rk050}1.169 & \cellcolor{rk050}0.520 & \cellcolor{rk050}0.127 & \cellcolor{rk100}0.736 & \cellcolor{rk050}15.01 & \cellcolor{rk100}1.012 & \cellcolor{rk100}2.189 \\
VGGT-$\Omega$ & Heuristic & \cellcolor{rk100}0.319 & \cellcolor{rk050}0.150 & \cellcolor{rk050}0.394 & \cellcolor{rk050}2.039 & \cellcolor{rk100}0.154 & \cellcolor{rk100}1.221 & \cellcolor{rk100}0.521 & \cellcolor{rk050}0.127 & \cellcolor{rk050}0.720 & \cellcolor{rk100}18.89 & \cellcolor{rk050}0.911 & \cellcolor{rk050}1.754 \\
\rowcolor{oursrow} & \method{} & \cellcolor{rk000}\best{0.179} & \cellcolor{rk000}\best{0.070} & \cellcolor{rk000}\best{0.352} & \cellcolor{rk000}\best{0.345} & \cellcolor{rk000}\best{0.073} & \cellcolor{rk000}\best{0.463} & \cellcolor{rk000}\best{0.153} & \cellcolor{rk000}\best{0.065} & \cellcolor{rk000}\best{0.556} & \cellcolor{rk000}\best{1.852} & \cellcolor{rk000}\best{0.140} & \cellcolor{rk000}\best{0.212} \\
\cmidrule(l){2-14}\addlinespace[0.3ex]
 & \textcolor{black!70}{\textit{no loop closure}} & \textcolor{black!70}{1.196} & \textcolor{black!70}{0.205} & \textcolor{black!70}{1.106} & \textcolor{black!70}{2.925} & \textcolor{black!70}{0.319} & \textcolor{black!70}{2.127} & \textcolor{black!70}{2.356} & \textcolor{black!70}{0.317} & \textcolor{black!70}{1.432} & \textcolor{black!70}{28.85} & \textcolor{black!70}{0.931} & \textcolor{black!70}{0.771} \\
 & Identity & \cellcolor{rk100}0.639 & \cellcolor{rk050}0.266 & \cellcolor{rk100}0.962 & \cellcolor{rk100}3.092 & \cellcolor{rk050}0.331 & \cellcolor{rk050}2.970 & \cellcolor{rk100}1.323 & \cellcolor{rk100}0.206 & \cellcolor{rk050}2.055 & \cellcolor{rk050}16.19 & \cellcolor{rk100}0.969 & \cellcolor{rk100}1.480 \\
VGGT & Heuristic & \cellcolor{rk050}0.638 & \cellcolor{rk100}0.269 & \cellcolor{rk050}0.946 & \cellcolor{rk050}3.085 & \cellcolor{rk050}0.331 & \cellcolor{rk100}2.978 & \cellcolor{rk050}1.267 & \cellcolor{rk050}0.196 & \cellcolor{rk100}2.144 & \cellcolor{rk100}16.39 & \cellcolor{rk050}0.913 & \cellcolor{rk050}1.402 \\
\rowcolor{oursrow} & \method{} & \cellcolor{rk000}\best{0.447} & \cellcolor{rk000}\best{0.158} & \cellcolor{rk000}\best{0.931} & \cellcolor{rk000}\best{1.630} & \cellcolor{rk000}\best{0.274} & \cellcolor{rk000}\best{2.123} & \cellcolor{rk000}\best{0.641} & \cellcolor{rk000}\best{0.190} & \cellcolor{rk000}\best{1.394} & \cellcolor{rk000}\best{10.02} & \cellcolor{rk000}\best{0.553} & \cellcolor{rk000}\best{0.775} \\
\cmidrule(l){2-14}\addlinespace[0.3ex]
 & \textcolor{black!70}{\textit{no loop closure}} & \textcolor{black!70}{0.394} & \textcolor{black!70}{0.096} & \textcolor{black!70}{0.556} & \textcolor{black!70}{1.215} & \textcolor{black!70}{0.117} & \textcolor{black!70}{0.703} & \textcolor{black!70}{1.255} & \textcolor{black!70}{0.166} & \textcolor{black!70}{1.161} & \textcolor{black!70}{8.166} & \textcolor{black!70}{0.277} & \textcolor{black!70}{0.448} \\
 & Identity & \cellcolor{rk100}0.443 & \cellcolor{rk100}0.242 & \cellcolor{rk050}0.571 & \cellcolor{rk050}1.858 & \cellcolor{rk050}0.188 & \cellcolor{rk050}1.259 & \cellcolor{rk050}1.138 & \cellcolor{rk050}0.158 & \cellcolor{rk050}1.860 & \cellcolor{rk100}19.98 & \cellcolor{rk100}0.928 & \cellcolor{rk100}2.799 \\
DA3 & Heuristic & \cellcolor{rk050}0.435 & \cellcolor{rk050}0.239 & \cellcolor{rk100}0.572 & \cellcolor{rk100}1.868 & \cellcolor{rk100}0.192 & \cellcolor{rk100}1.365 & \cellcolor{rk100}1.145 & \cellcolor{rk100}0.159 & \cellcolor{rk100}1.952 & \cellcolor{rk050}18.71 & \cellcolor{rk050}0.759 & \cellcolor{rk050}2.162 \\
\rowcolor{oursrow} & \method{} & \cellcolor{rk000}\best{0.226} & \cellcolor{rk000}\best{0.090} & \cellcolor{rk000}\best{0.538} & \cellcolor{rk000}\best{0.387} & \cellcolor{rk000}\best{0.094} & \cellcolor{rk000}\best{0.677} & \cellcolor{rk000}\best{0.242} & \cellcolor{rk000}\best{0.086} & \cellcolor{rk000}\best{1.034} & \cellcolor{rk000}\best{2.023} & \cellcolor{rk000}\best{0.158} & \cellcolor{rk000}\best{0.447} \\
\midrule
\multicolumn{2}{@{}l}{\textit{(c) Visual-inertial initialization}} & \metric{$e_s$} & \metric{$e_g$} & \metric{$e_{b_g}$} & \metric{$e_s$} & \metric{$e_g$} & \metric{$e_{b_g}$} & \metric{$e_s$} & \metric{$e_g$} & \metric{$e_{b_g}$} & \metric{$e_s$} & \metric{$e_g$} & \metric{$e_{b_g}$} \\
\cmidrule(r){1-2}\cmidrule(lr){3-5}\cmidrule(lr){6-8}\cmidrule(lr){9-11}\cmidrule(lr){12-14}
 & Identity & \cellcolor{rk100}18.6 & \cellcolor{rk050}0.86 & \cellcolor{rk050}0.169 & \cellcolor{rk050}12.2 & \cellcolor{rk050}1.09 & \cellcolor{rk050}0.262 & \cellcolor{rk100}9.56 & \cellcolor{rk050}0.48 & \cellcolor{rk050}0.295 & \cellcolor{rk050}31.9 & \cellcolor{rk050}0.76 & \cellcolor{rk050}0.100 \\
VGGT-$\Omega$ & Heuristic & \cellcolor{rk050}15.8 & \cellcolor{rk050}0.86 & \cellcolor{rk050}0.169 & \cellcolor{rk100}12.3 & \cellcolor{rk050}1.09 & \cellcolor{rk050}0.262 & \cellcolor{rk050}9.28 & \cellcolor{rk050}0.48 & \cellcolor{rk050}0.295 & \cellcolor{rk100}32.0 & \cellcolor{rk100}0.77 & \cellcolor{rk050}0.100 \\
\rowcolor{oursrow} & \method{} & \cellcolor{rk000}\best{11.4} & \cellcolor{rk000}\best{0.85} & \cellcolor{rk000}\best{0.153} & \cellcolor{rk000}\best{10.0} & \cellcolor{rk000}\best{1.06} & \cellcolor{rk000}\best{0.220} & \cellcolor{rk000}\best{6.17} & \cellcolor{rk000}\best{0.41} & \cellcolor{rk000}\best{0.216} & \cellcolor{rk000}\best{30.9} & \cellcolor{rk000}\best{0.74} & \cellcolor{rk000}\best{0.096} \\
\cmidrule(l){2-14}
 & Identity & \cellcolor{rk100}32.8 & \cellcolor{rk050}1.26 & \cellcolor{rk050}0.588 & \cellcolor{rk050}31.2 & \cellcolor{rk000}\best{1.58} & \cellcolor{rk050}1.108 & \cellcolor{rk100}30.3 & \cellcolor{rk100}0.92 & \cellcolor{rk050}0.715 & \cellcolor{rk100}43.7 & \cellcolor{rk050}0.86 & \cellcolor{rk050}0.391 \\
VGGT & Heuristic & \cellcolor{rk050}29.8 & \cellcolor{rk100}1.27 & \cellcolor{rk050}0.588 & \cellcolor{rk050}31.2 & \cellcolor{rk000}\best{1.58} & \cellcolor{rk050}1.108 & \cellcolor{rk050}28.5 & \cellcolor{rk050}0.91 & \cellcolor{rk050}0.715 & \cellcolor{rk050}43.4 & \cellcolor{rk100}0.87 & \cellcolor{rk050}0.391 \\
\rowcolor{oursrow} & \method{} & \cellcolor{rk000}\best{19.4} & \cellcolor{rk000}\best{1.24} & \cellcolor{rk000}\best{0.493} & \cellcolor{rk000}\best{23.3} & \cellcolor{rk000}\best{1.58} & \cellcolor{rk000}\best{0.768} & \cellcolor{rk000}\best{17.1} & \cellcolor{rk000}\best{0.76} & \cellcolor{rk000}\best{0.491} & \cellcolor{rk000}\best{39.6} & \cellcolor{rk000}\best{0.77} & \cellcolor{rk000}\best{0.278} \\
\cmidrule(l){2-14}
 & Identity & \cellcolor{rk100}16.1 & \cellcolor{rk000}\best{0.95} & \cellcolor{rk050}0.226 & \cellcolor{rk050}15.5 & \cellcolor{rk000}\best{1.15} & \cellcolor{rk050}0.355 & \cellcolor{rk100}11.1 & \cellcolor{rk050}0.58 & \cellcolor{rk050}0.577 & \cellcolor{rk050}32.9 & \cellcolor{rk050}0.80 & \cellcolor{rk050}0.211 \\
DA3 & Heuristic & \cellcolor{rk050}15.1 & \cellcolor{rk100}0.96 & \cellcolor{rk050}0.226 & \cellcolor{rk100}15.6 & \cellcolor{rk000}\best{1.15} & \cellcolor{rk050}0.355 & \cellcolor{rk050}11.0 & \cellcolor{rk050}0.58 & \cellcolor{rk050}0.577 & \cellcolor{rk100}33.2 & \cellcolor{rk050}0.80 & \cellcolor{rk050}0.211 \\
\rowcolor{oursrow} & \method{} & \cellcolor{rk000}\best{13.1} & \cellcolor{rk000}\best{0.95} & \cellcolor{rk000}\best{0.218} & \cellcolor{rk000}\best{14.8} & \cellcolor{rk000}\best{1.15} & \cellcolor{rk000}\best{0.295} & \cellcolor{rk000}\best{9.22} & \cellcolor{rk000}\best{0.57} & \cellcolor{rk000}\best{0.408} & \cellcolor{rk000}\best{29.6} & \cellcolor{rk000}\best{0.70} & \cellcolor{rk000}\best{0.183} \\
\bottomrule
\end{tabular}}\global\isotabwidth=\wd\isotabbox %
\usebox{\isotabbox}
\label{tab:isolated}
\end{table*}

\begin{table*}[t]
\centering
\footnotesize
\setlength{\tabcolsep}{1.5pt}
\caption{
\textbf{Complete foundation-model-based systems.}
We evaluate \method{} in VGGT-Bridge and FoundationVIO using VGGT-$\Omega$,
VGGT, and DA3, without (w/o) and with (w/) loop closure.
\fail{} indicates divergence on at least one sequence.
Best per backbone and setting in \textbf{bold}.}
\begin{tabular*}{\isotabwidth}{@{\extracolsep{\fill}}llcccccccccccccccc@{}}%
\toprule
 & & \multicolumn{8}{c}{VGGT-Bridge~\cite{choi2026vggtbridge} (Sim(3) ATE, m)} & \multicolumn{8}{c}{FoundationVIO~\cite{rhee2026foundationvio} (SE(3) ATE, m)} \\
\cmidrule(lr){3-10}\cmidrule(l){11-18}\addlinespace[0.3ex]
 & & \multicolumn{2}{c}{EuRoC} & \multicolumn{2}{c}{Aria} & \multicolumn{2}{c}{ScaRF} & \multicolumn{2}{c}{Spires} & \multicolumn{2}{c}{EuRoC} & \multicolumn{2}{c}{Aria} & \multicolumn{2}{c}{ScaRF} & \multicolumn{2}{c}{Spires} \\
\cmidrule(lr){3-4}\cmidrule(lr){5-6}\cmidrule(lr){7-8}\cmidrule(lr){9-10}\cmidrule(lr){11-12}\cmidrule(lr){13-14}\cmidrule(lr){15-16}\cmidrule(lr){17-18}\addlinespace[0.3ex]
Backbone & Weighting & \metric{w/o} & \metric{w/} & \metric{w/o} & \metric{w/} & \metric{w/o} & \metric{w/} & \metric{w/o} & \metric{w/} & \metric{w/o} & \metric{w/} & \metric{w/o} & \metric{w/} & \metric{w/o} & \metric{w/} & \metric{w/o} & \metric{w/} \\
\midrule
VGGT-$\Omega$ & Default & \cellcolor{rk100}0.204 & \cellcolor{rk100}0.113 & \cellcolor{rk100}0.624 & \cellcolor{rk100}0.381 & \cellcolor{rk100}1.546 & \cellcolor{rk100}0.638 & \cellcolor{rk000}\best{3.783} & \cellcolor{rk100}1.915 & \cellcolor{rk100}0.663 & \cellcolor{rk100}0.267 & \cellcolor{rk100}0.568 & \cellcolor{rk100}0.372 & \cellcolor{rk100}0.784 & \cellcolor{rk100}0.365 & \cellcolor{rk100}28.42 & \cellcolor{rk100}28.71 \\
\rowcolor{oursrow} & \method{} & \cellcolor{rk000}\best{0.140} & \cellcolor{rk000}\best{0.073} & \cellcolor{rk000}\best{0.368} & \cellcolor{rk000}\best{0.178} & \cellcolor{rk000}\best{1.155} & \cellcolor{rk000}\best{0.455} & \cellcolor{rk100}4.442 & \cellcolor{rk000}\best{0.792} & \cellcolor{rk000}\best{0.287} & \cellcolor{rk000}\best{0.121} & \cellcolor{rk000}\best{0.533} & \cellcolor{rk000}\best{0.258} & \cellcolor{rk000}\best{0.348} & \cellcolor{rk000}\best{0.125} & \cellcolor{rk000}\best{5.427} & \cellcolor{rk000}\best{5.445} \\
\cmidrule(l){2-18}
VGGT & Default & \cellcolor{rk100}1.099 & \cellcolor{rk100}0.632 & \cellcolor{rk100}3.903 & \cellcolor{rk100}3.039 & \cellcolor{rk100}4.137 & \cellcolor{rk100}3.501 & \cellcolor{rk100}17.77 & \cellcolor{rk100}14.66 & \cellcolor{rk100}1.090 & \cellcolor{rk100}0.427 & \cellcolor{rk100}2.190 & \cellcolor{rk100}1.258 & \cellcolor{rk100}1.298 & \cellcolor{rk100}0.511 & \cellcolor{rk100}\fail & \cellcolor{rk100}\fail \\
\rowcolor{oursrow} & \method{} & \cellcolor{rk000}\best{0.912} & \cellcolor{rk000}\best{0.526} & \cellcolor{rk000}\best{3.677} & \cellcolor{rk000}\best{2.408} & \cellcolor{rk000}\best{3.607} & \cellcolor{rk000}\best{2.833} & \cellcolor{rk000}\best{11.28} & \cellcolor{rk000}\best{5.264} & \cellcolor{rk000}\best{0.767} & \cellcolor{rk000}\best{0.309} & \cellcolor{rk000}\best{1.492} & \cellcolor{rk000}\best{0.951} & \cellcolor{rk000}\best{0.928} & \cellcolor{rk000}\best{0.209} & \cellcolor{rk100}\fail & \cellcolor{rk100}\fail \\
\cmidrule(l){2-18}
DA3 & Default & \cellcolor{rk100}0.347 & \cellcolor{rk100}0.229 & \cellcolor{rk100}1.773 & \cellcolor{rk100}1.017 & \cellcolor{rk100}1.850 & \cellcolor{rk100}0.921 & \cellcolor{rk100}4.610 & \cellcolor{rk100}3.531 & \cellcolor{rk100}0.724 & \cellcolor{rk100}0.363 & \cellcolor{rk100}0.704 & \cellcolor{rk100}0.398 & \cellcolor{rk100}0.955 & \cellcolor{rk100}0.355 & \cellcolor{rk100}28.68 & \cellcolor{rk100}30.17 \\
\rowcolor{oursrow} & \method{} & \cellcolor{rk000}\best{0.244} & \cellcolor{rk000}\best{0.133} & \cellcolor{rk000}\best{1.388} & \cellcolor{rk000}\best{0.790} & \cellcolor{rk000}\best{1.418} & \cellcolor{rk000}\best{0.484} & \cellcolor{rk000}\best{3.615} & \cellcolor{rk000}\best{1.905} & \cellcolor{rk000}\best{0.389} & \cellcolor{rk000}\best{0.194} & \cellcolor{rk000}\best{0.496} & \cellcolor{rk000}\best{0.315} & \cellcolor{rk000}\best{0.685} & \cellcolor{rk000}\best{0.163} & \cellcolor{rk000}\best{3.600} & \cellcolor{rk000}\best{3.232} \\
\bottomrule
\end{tabular*}
\label{tab:systems}
\end{table*}

\begin{figure*}[t]
    \centering
    \includegraphics[width=0.8\textwidth]{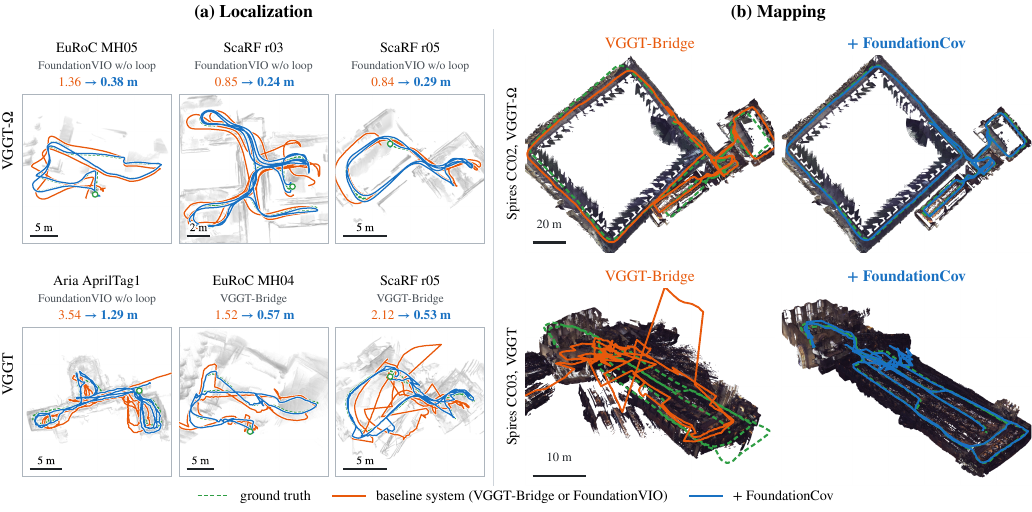}
    \caption{\textbf{Qualitative results of the complete systems.}
    (a)~Localization: trajectories of VGGT-Bridge or FoundationVIO with the default weights (orange) and with \method{} (blue) against the ground truth (green, dashed), over the floor plan (gray);
    above each plot the ATE RMSE (m).
    (b)~Mapping: the maps built by the same systems, rendered from the same viewpoint for both weightings.}
    \label{fig:qualitative}
\end{figure*}

\subsection{Isolated Estimation Tasks}
\label{sec:exp_isolated}

GFMs are commonly used in estimation systems in three ways: aligning overlapping windows~\cite{deng2026vggt,maggio2026vggt,choi2026vggtbridge}, closing loops in a pose graph~\cite{zhang2025vistaslam,xiong2026vggtmotion}, and initializing visual--inertial estimation~\cite{rhee2026foundationvio,zhang2026ffvioinit}.
We study each in isolation, changing only the weighting of the GFM predictions.
In every task we compare three weightings of the same estimator: \emph{Identity} (unit covariances), \emph{Heuristic} (the confidence-based or constant weighting of existing systems), and \method{} (our predicted covariances).

\paragraph{Sim(3) alignment.}
Two overlapping windows are reconstructed by separate GFM inferences, each in its own coordinate frame and scale.
We estimate the Sim(3) transformation $S$ between them from the 3D points that both predict for their shared frames.
With windows of 30 frames overlapping by 15 (dataset-specific settings in \cref{sec:supp_isolated}), we solve
\begin{equation}
  \hat S = \arg\min_S \sum_p
  \big(\mathbf{x}^{(1)}_p-S\mathbf{x}^{(2)}_p\big)^{\top}
  \mathbf{W}_p
  \big(\mathbf{x}^{(1)}_p-S\mathbf{x}^{(2)}_p\big),
  \label{eq:exp_sim3_alignment}
\end{equation}
where $\mathbf{x}^{(w)}_p$ is the 3D point predicted for pixel $p$ by window $w$ and the sum runs over all shared pixels, without pixel selection or a robust kernel.
Identity sets $\mathbf{W}_p=\mathbf{I}$ (Umeyama), Heuristic sets $\mathbf{W}_p=\sqrt{c^{(1)}_p c^{(2)}_p}\,\mathbf{I}$ from the two predicted confidences, following VGGT-Long~\cite{deng2026vggt}, and \method{} sets $\mathbf{W}_p=\boldsymbol{\Sigma}_p^{-1}$, the inverse covariance of the residual, propagated from the predicted depth, pose, and focal-length covariances (\cref{sec:supp_propagation}).

\paragraph{Loop closure.}
Loop closure corrects accumulated drift, and its result depends on how the loop constraints are weighted against the sequential ones.
We study this dependence in a pose graph built only from GFM predictions.
The sequential constraints are the relative poses that the GFM predicts between consecutive frames, on 10-frame windows that overlap by one frame; scale factors link the scales of neighboring windows (factor definitions in \cref{sec:supp_isolated}).
Each revisit found by image retrieval adds a loop constraint: the relative pose that the GFM predicts for the two frames as a pair.
Without loop constraints the graph has no redundancy, so every weighting gives the same open-loop trajectory.
For all relative pose factors, Identity uses unit covariances, Heuristic uses $\boldsymbol{\Sigma}_{ij}=\mathbf{I}/w_{ij}$, where $w_{ij}$ is the mean confidence of the two frames normalized by its median over the sequence, and \method{} uses the predicted $6\times6$ covariance of the frame pair.

\paragraph{Visual-inertial initialization.}
We evaluate loosely coupled visual--inertial initialization following VINS-Mono~\cite{qin2018vins} and FoundationVIO~\cite{rhee2026foundationvio}: a linear solver aligns IMU preintegration with the GFM's relative poses over two overlapping windows to estimate the gyroscope bias, gravity direction, and metric scale.
The weightings change only the noise model of this solver: Identity is the released unweighted solver, Heuristic uses FoundationVIO's noise model
(the IMU noise of each dataset's calibration plus a constant pose noise of $4^\circ$ and 0.01\,m), and \method{} replaces the constant pose noise by the predicted pose covariance.
Following~\cite{fei2026maci2,zhang2026ffvioinit}, we initialize the 1637 non-overlapping windows passing the released excitation test and report
\begin{equation}
  e_s=100\,\Big|\frac{\hat s}{s^{\mathrm{gt}}}-1\Big|,\quad
  e_g=\angle\big(\hat{\mathbf{g}},\mathbf{g}^{\mathrm{gt}}\big),\quad
  e_{b_g}=\big\|\hat{\mathbf{b}}_g-\mathbf{b}_g^{\mathrm{ref}}\big\|,
  \label{eq:exp_vi}
\end{equation}
where $s^{\mathrm{gt}}$ is the Umeyama scale of the window against ground truth and $\mathbf{b}_g^{\mathrm{ref}}$ the bias estimated from the ground-truth rotations over $\pm10$\,s.

\paragraph{Results.}
Tab.~\ref{tab:isolated} shows gains across all three tasks and GFMs.
For Sim(3) alignment, \method{} gives the lowest rotation and translation errors on every dataset, reducing translation error by 11--61\% over Heuristic.
For loop closure, it reduces mean ATE by 43--85\% from the common open-loop trajectory, including cases where Identity and Heuristic worsen it.
For visual--inertial initialization, scale, bias, and gravity-direction errors decrease by up to 40\%, 31\%, and 16\% over Heuristic, respectively.

\subsection{Complete Systems}
\label{sec:exp_systems}

We apply \method{} to a visual and a visual--inertial SLAM system built on GFMs by replacing the weights in their optimization problems with our predicted covariances.

\paragraph{VGGT-Bridge.}
VGGT-Bridge~\cite{choi2026vggtbridge} builds a Sim(3) pose graph over
chunks at three temporal strides, with sequential, loop, and skip edges
obtained by point-cloud alignment.
The system uses unweighted alignment and sets edge weights to 1,
or to $0.3\min(1,n/20000)$ for a skip edge with $n$ points.
With \method{}, we compute each edge using the weighted alignment in
Eq.~\eqref{eq:exp_sim3_alignment}, with
$\mathbf{W}_p=\boldsymbol{\Sigma}_p^{-1}$.
VGGT-Bridge back-projects all frames using shared reference intrinsics,
after rescaling each frame's depth by the mean ratio of the reference
to its predicted focal lengths.
We propagate the predicted focal-length covariances through this
rescaling and back-projection when computing $\boldsymbol{\Sigma}_p$.
Each edge is then weighted using its first-order covariance.
Chunking, graph connectivity, and the optimizer are unchanged.

\paragraph{FoundationVIO.}
FoundationVIO~\cite{rhee2026foundationvio} fuses the relative poses predicted on batches of 10 frames with IMU preintegration in iSAM2, with constant noise on the relative pose factors.
\method{} replaces this noise by the predicted covariance of each frame pair, mapped into the factor residual; the scale factors and the Huber kernel on loop factors are unchanged.
Since the visual factors now carry predicted covariances, all rows use the IMU noise of each dataset's own calibration rather than values tuned per dataset.

\paragraph{Results.}
Tab.~\ref{tab:systems} shows that \method{} achieves substantial gains in accuracy in both complete systems, VGGT-Bridge and
FoundationVIO, across all GFMs.
\emph{With loop closure}, it reduces ATE by 17--64\% in VGGT-Bridge and 21--89\%
in FoundationVIO relative to Default.
The largest improvement is obtained by FoundationVIO with DA3 on Spires,
where ATE decreases from 30.17 to 3.23\,m.
\emph{Without loop closure}, \method{} also improves both systems, with
ATE reductions of up to 41\% in VGGT-Bridge and 87\% in FoundationVIO
relative to Default.
Spires is the only dataset where FoundationVIO fails with VGGT under both
weightings, which we attribute to VGGT's larger relative-rotation errors.
Qualitative results in Fig.~\ref{fig:qualitative} further show that \method{} brings the estimated trajectories into closer agreement with the ground truth and reduces structural distortions in the reconstructed maps.
\subsection{Which Features Predict Covariance?}
\label{sec:input_ablation}
\begin{table}[t]
\centering
\scriptsize
\setlength{\tabcolsep}{1.1pt}
\caption{\textbf{Input ablation.} Averages on the datasets in \cref{tab:calibration}.}
\label{tab:input_ablation}
\begin{tabular*}{\columnwidth}{@{\extracolsep{\fill}}l rr@{\hspace{5pt}}rr@{\hspace{5pt}}rr@{}}
\toprule
& \multicolumn{2}{c}{VGGT-$\Omega$} & \multicolumn{2}{c}{VGGT} & \multicolumn{2}{c}{DA3} \\
\cmidrule(lr){2-3}\cmidrule(lr){4-5}\cmidrule(lr){6-7}\addlinespace[0.3ex]
Input & AUCE$\downarrow$ & NLL$\downarrow$ & AUCE$\downarrow$ & NLL$\downarrow$ & AUCE$\downarrow$ & NLL$\downarrow$ \\
\midrule
\multicolumn{7}{@{}l}{\textit{Relative pose}} \\
Predictions only & \cellcolor{rk100}0.408 & \cellcolor{rk100}-12.26 & \cellcolor{rk100}0.270 & \cellcolor{rk100}-11.07 & \cellcolor{rk100}0.356 & \cellcolor{rk100}-10.53 \\
+ Image & \cellcolor{rk050}0.260 & \cellcolor{rk050}-17.07 & \cellcolor{rk050}0.189 & \cellcolor{rk050}-12.00 & \cellcolor{rk050}0.241 & \cellcolor{rk050}-12.33 \\
+ Geometric & \cellcolor{rk000}\textbf{0.069} & \cellcolor{rk000}\textbf{-18.18} & \cellcolor{rk000}\textbf{0.110} & \cellcolor{rk000}\textbf{-13.89} & \cellcolor{rk000}\textbf{0.153} & \cellcolor{rk000}\textbf{-15.15} \\
\midrule
\multicolumn{7}{@{}l}{\textit{Focal length}} \\
Predictions only & \cellcolor{rk100}0.224 & \cellcolor{rk100}-5.31 & \cellcolor{rk100}0.233 & \cellcolor{rk100}-3.98 & \cellcolor{rk100}0.221 & \cellcolor{rk100}-3.48 \\
+ Image & \cellcolor{rk050}0.162 & \cellcolor{rk000}\textbf{-5.70} & \cellcolor{rk050}0.161 & \cellcolor{rk000}\textbf{-4.81} & \cellcolor{rk050}0.196 & \cellcolor{rk050}-3.70 \\
+ Geometric & \cellcolor{rk000}\textbf{0.087} & \cellcolor{rk050}-5.69 & \cellcolor{rk000}\textbf{0.029} & \cellcolor{rk050}-4.77 & \cellcolor{rk000}\textbf{0.102} & \cellcolor{rk000}\textbf{-4.58} \\
\midrule
\multicolumn{7}{@{}l}{\textit{Depth}} \\
Depth + conf. & \cellcolor{rk000}\textbf{0.089} & \cellcolor{rk067}-1.52 & \cellcolor{rk033}0.255 & \cellcolor{rk067}-0.25 & \cellcolor{rk033}0.200 & \cellcolor{rk033}-0.25 \\
Dense features only & \cellcolor{rk033}0.097 & \cellcolor{rk033}-1.53 & \cellcolor{rk067}0.258 & \cellcolor{rk033}-0.26 & \cellcolor{rk067}0.208 & \cellcolor{rk067}-0.22 \\
+ Image & \cellcolor{rk100}0.135 & \cellcolor{rk100}-1.49 & \cellcolor{rk100}0.261 & \cellcolor{rk100}0.18 & \cellcolor{rk100}0.263 & \cellcolor{rk100}0.21 \\
+ Camera-head & \cellcolor{rk067}0.099 & \cellcolor{rk000}\textbf{-1.60} & \cellcolor{rk000}\textbf{0.165} & \cellcolor{rk000}\textbf{-0.65} & \cellcolor{rk000}\textbf{0.114} & \cellcolor{rk000}\textbf{-0.42} \\
\bottomrule
\end{tabular*}
\end{table}

\method{} rests on the premise that the features a GFM computes for reconstruction also carry information about its errors. We test this premise by changing only the input of each head, training each variant for 6k steps.
For pose and focal length, we compare the GFM's \emph{predictions only}; \emph{+ image}, which adds tokens of the per-image encoder (DINO; the pre-camera CLS token for DA3); and \emph{+ geometric}, which adds the tokens after cross-view attention that our heads use.
This tests whether features that combine information across views
predict covariances better than features from single images.

For pose, geometric tokens are best for all three GFMs in both metrics (\cref{tab:input_ablation}): they lower AUCE by a further 37--73\% over image tokens, which themselves improve on predictions alone.
For focal length, they lower AUCE by 46--82\% relative to image tokens, at comparable or better NLL.
For depth, adding camera-head features to the dense features (\cref{eq:depth_head}) improves NLL for all three GFMs and AUCE for VGGT and DA3, whereas adding image tokens instead degrades both.
A pointwise head on predicted depth and native confidence alone is competitive in AUCE for VGGT-$\Omega$ but worse in NLL for every GFM.

In all three heads, the most useful inputs are those the GFM computes after fusing the views, which supports predicting covariances from its frozen representation.

\section{Discussion and Limitations}
\label{sec:discussion}
\method{} asks little of the GFM: frozen camera tokens for the pose and focal heads and dense decoder features for the depth head, with one head architecture for VGGT, VGGT-$\Omega$, and Depth Anything 3.
We therefore expect it to carry over to other GFMs that expose such features, although we have not tested any beyond these three, in particular models that predict pointmaps without explicit camera poses.
\method{} has two limitations: (i) its distributions are zero-mean Gaussian marginals, without correlations across pixels or frame pairs or between depth, focal length, and pose; and (ii) their scale is accurate only up to a factor of about 1.6 on unseen data.
Future work can target on these issues.

\section{Conclusion}
We presented \method{}, which equips frozen GFMs with predictive covariances for depth, relative pose, and focal length without altering their predictions.
Across three GFMs, these covariances improve isolated estimation tasks and complete SLAM systems, turning GFMs into probabilistic front ends for downstream estimation.

\section*{Acknowledgments}
We thank AIST for access to the ABCI computing cluster.
This research was conducted in part under FRONTia, a Japanese national program led by METI and NEDO.

{
    \small
    \bibliographystyle{ieeenat_fullname}
    \bibliography{main}
}

\clearpage
\setcounter{section}{0}
\setcounter{subsection}{0}
\setcounter{subsubsection}{0}
\setcounter{figure}{0}
\setcounter{table}{0}
\setcounter{equation}{0}

\renewcommand{\thesection}{S\arabic{section}}
\renewcommand{\thefigure}{S\arabic{figure}}
\renewcommand{\thetable}{S\arabic{table}}
\renewcommand{\theequation}{S\arabic{equation}}

\renewcommand{\theHsection}{supp.\arabic{section}}
\renewcommand{\theHfigure}{supp.\arabic{figure}}
\renewcommand{\theHtable}{supp.\arabic{table}}
\renewcommand{\theHequation}{supp.\arabic{equation}}
\maketitlesupplementary

This supplement provides implementation and training details (\cref{sec:supp_method}), covariance propagation formulas (\cref{sec:supp_propagation}), and the objectives and weighting schemes used in downstream experiments (\cref{sec:supp_downstream}).
It also reports additional ablations (\cref{sec:supp_additional_ablations}), further experiments (\cref{sec:supp_additional}), and per-sequence results (\cref{sec:supp_perseq}).

\section{Architecture and Training}
\label{sec:supp_method}

\subsection{Feature Sources and Prediction Heads}

\paragraph{Pose.}
For predicted relative pose $(\hat R_{ij},\hat t_{ij})$, the pair descriptor concatenates the two projected frame features $c_i,c_j\in\mathbb R^{256}$ with
\[
 [\hat t_{ij},\ (\hat R_{ij})_{:,1},\ (\hat R_{ij})_{:,2},\ k_i,\ k_j]\in\mathbb R^{17},
\]
where $k_i$ contains the two normalized focal lengths and principal-point coordinates. Each frame feature uses the last transformer layer's camera token and, where available, the mean register token (16 registers for VGGT-$\Omega$, four for VGGT, none for DA3). A projection of the pair descriptor and an embedding of log translation magnitude and log rotation angle are added before two residual MLP blocks.

\paragraph{Depth.}
A trainable copy of the pretrained DPT decodes patch features from transformer layers $(5,12,18,24)$ for VGGT-$\Omega$ and VGGT, or $(12,16,20,24)$ for DA3, counting the first layer as 1. The original depth decoder stays frozen. At each of $128\times128$ locations, a pointwise MLP reads the decoded feature and a projected frame feature from the camera head. For scalar output $a$, the normalized log variance is $\ell^d=16\tanh(a/16)$ and the additive depth variance is $\hat d^2\exp(\ell^d)$.

\paragraph{Focal length.}
Let $\hat\kappa_i$ be the two predicted log focal lengths defined in \cref{sec:method_errors}. The six window statistics in \cref{eq:focal_head} are
\[
 \begin{aligned}
 m_i &= \operatorname{median}_{j\ne i}\hat\kappa_j, &
 q_i &= \operatorname{median}_{j\ne i}|\hat\kappa_j-m_i|,\\
 a_i &= [m_i,\ q_i,\ \hat\kappa_i-m_i].
 \end{aligned}
\]
Medians and absolute values act coordinatewise. The head sums learned projections of $a_i$, $\hat\kappa_i$, and the final transformer camera/register feature. It predicts a lower-triangular factor with softplus on the diagonal and uses a covariance floor $\epsilon_K^2 I$ with $\epsilon_K=10^{-3}$. Focal supervision uses windows with at least two valid frames sharing the same pinhole calibration after preprocessing.

\subsection{Pose Covariance Parameterization}

The pose head predicts 21 values: six log-variance offsets $a\in\mathbb R^6$ and the fifteen strictly lower-triangular entries of a unit-diagonal matrix $B$.
We set $\ell=2\log(0.1)\mathbf 1+a$ and construct
\begin{equation}
 \begin{aligned}
 G&=BB^\top,\qquad Q=\operatorname{diag}(G_{11},\ldots,G_{66}),\\
 C&=(1-\lambda)Q^{-1/2}GQ^{-1/2}+\lambda I_6,\\
 S&=\operatorname{diag}(\exp(\ell/2)),\qquad \Sigma=SCS,
 \end{aligned}
 \label{eq:supp_pose_parameterization}
\end{equation}
with $\lambda=10^{-4}$.
The unit diagonal of $C$ gives $\Sigma_{kk}=\exp(\ell_k)$, while the identity mixture keeps $C$ positive definite. For $i<j$, the coordinates are local translation followed by rotation. Reversing the pair gives $\Sigma_{ji}=\operatorname{Ad}_{\hat T_{ij}}\Sigma_{ij}\operatorname{Ad}_{\hat T_{ij}}^\top$ to first order.

\subsection{Training Objective and Optimization}

Let $\hat p_i$ and $p_i$ be the predicted and ground-truth camera positions, each expressed in its first camera's frame. The least-squares scale $s=(\sum_i\hat p_i^\top p_i)/(\sum_i\|\hat p_i\|^2)$ converts reconstruction units to ground-truth units. For windows with an identifiable positive scale, we divide ground-truth positions and depths by $s$ to form the targets in \cref{sec:method_errors}. At inference the heads return covariance directly in reconstruction units.

With Gaussian NLL $\ell$ from \cref{eq:gaussian_nll}, the objective is
\[
 \mathcal L=\mathcal L_{\rm pose}+0.2\,\mathcal L_{\rm depth}+\mathcal L_{\rm focal}.
\]
Each term uses its available ground truth. Non-adjacent pose pairs receive half the weight of adjacent pairs. For depth, we average valid pixels within frames, then valid frames within windows. At a pixel with $\hat d>0$,
\[
 \ell(e^d,\hat d^2 v)=\ell(e^d/\hat d,v)+\log\hat d .
\]
We train with the normalized term on the right. Both forms give the same gradients because the predicted depth is frozen.

We sample the 16 datasets equally. AdamW uses $\beta=(0.9,0.95)$ and weight decay $0.05$. Over 17,000 updates, the learning rate rises linearly from $5\times10^{-7}$ to $2\times10^{-4}$ in the first 5\%, then decays by a cosine schedule to $5\times10^{-7}$.

\section{Covariance Propagation}
\label{sec:supp_propagation}
\label{sec:supp_syscov}

\subsection{Point Covariance}

For pixel $p$ in frame $m$, let $x^c_p=\hat d_p\hat K_m^{-1}\tilde u_p$ be its back-projection in camera coordinates and $x_p=R_m x^c_p+t_m$ its position in the first camera's frame. Write $x^c_p=(X_p,Y_p,Z_p)^\top$ and $a_p=R_m x^c_p$. A depth error moves this point along $a_p$, while a pose or focal error moves all points from frame $m$ together. Assuming independence between the three error sources, the marginal point covariance is
\begin{equation}
 \begin{aligned}
  \Sigma_p &= \Sigma^d_p + J_p\Sigma_{\eta,m}J_p^\top
                      + B_p\Sigma^{\rm focal}_m B_p^\top,\\
  \Sigma^d_p &= \exp(\ell^d_p)\,a_pa_p^\top,\qquad
  J_p=[I_3,\ -[x_p]_\times].
 \end{aligned}
 \label{eq:supp_point_cov}
\end{equation}
Here $[x]_\times y=x\times y$. The pose covariance in window coordinates is $\Sigma_{\eta,m}=\operatorname{Ad}_{\hat T_{1m}}\Sigma^{\rm pose}_{1m}\operatorname{Ad}_{\hat T_{1m}}^\top$, obtained from the local perturbation of \cref{eq:pose_error}.

The focal Jacobian $\mathbf{B}_p$ depends on the back-projection procedure.
In isolated alignment, each frame uses its own predicted intrinsics, giving $\mathbf{B}_p=\mathbf{R}_m\big[\,X_p\mathbf{u}_1,\ Y_p\mathbf{u}_2\big]$, where $\mathbf{u}_1$ and $\mathbf{u}_2$ are coordinate unit vectors.
VGGT-Bridge groups overlapping reconstruction windows (chunks) by temporal stride. Each group shares one reference calibration. The system rescales each frame's depth by the mean of the reference-to-predicted focal ratios $\mathbf{r}_m$.
Its focal Jacobian is therefore $\mathbf{B}_p=\mathbf{a}_p\mathbf{w}_m^{\top}$, where $\mathbf{w}_m=\mathbf{r}_m/(\mathbf{1}^{\top}\mathbf{r}_m)$.
Reference-focal error cancels within a chunk set.
For a skip edge linking chunks at different strides, it contributes the additional shared term $\mathbf{Q}_p\big(\boldsymbol{\Sigma}^{\mathrm{focal}}_{\mathrm{ref},1}+\boldsymbol{\Sigma}^{\mathrm{focal}}_{\mathrm{ref},2}\big)\mathbf{Q}_p^{\top}$, with
\begin{equation}
  \mathbf{Q}_p=\tfrac12\,\mathbf{R}_m\begin{bmatrix}-X_p & X_p\\ Y_p & -Y_p\\ Z_p & Z_p\end{bmatrix}.
\end{equation}
\subsection{Sim(3) Registration Covariance}
For corresponding points $x_p^{(1)},x_p^{(2)}$, let $S=(s,R,t)$ map window 2 into window 1. Under independence between windows, the point residual and its covariance are
\begin{equation}
 \begin{aligned}
 e_p(S)&=sR x_p^{(2)}+t-x_p^{(1)},\\
 C_p(S)&=\Sigma_p^{(1)}+s^2R\Sigma_p^{(2)}R^\top .
 \end{aligned}
 \label{eq:alignment_covariance}
\end{equation}
Registration weights each point by $W_p=C_p(S)^{-1}$, updated with the current scale and rotation.

\paragraph{Uncertainty of the estimated transformation.}
For the right perturbation $\hat S=S\,\mathrm{Exp}(\boldsymbol{\xi})$, $\boldsymbol{\xi}=(\boldsymbol{\rho},\boldsymbol{\phi},\log\sigma)$, the Jacobian of $\mathbf{e}_p$ is $\mathbf{G}_p=sR\,[\,\mathbf{I},\ -[\mathbf{x}^{(2)}_p]_\times,\ \mathbf{x}^{(2)}_p\,]$.
At the solution, fix the weights and let $U_p=G_p^\top W_p$ and $H=\sum_p U_pG_p$. A perturbation $\delta_p$ of the point residual then changes the fitted transformation by $\xi\approx-H^{-1}\sum_p U_p\delta_p$.

Depth errors are modeled independently across pixels. Pose and focal errors are shared across pixels of a frame and modeled independently across frames and windows. This gives
\begin{equation}
 \begin{aligned}
 \Sigma_\xi &= H^{-1}(V_d+V_T+V_K)H^{-1},\\
 V_d &= \sum_p U_p C^d_p U_p^\top,\\
 V_T &= \sum_{w,m} G_{w,m}\Sigma_{\eta,w,m}G_{w,m}^\top,\\
 V_K &= \sum_{w,m} G^K_{w,m}\Sigma^{\rm focal}_{w,m}(G^K_{w,m})^\top .
 \end{aligned}
 \label{eq:exp_alignment_cov}
\end{equation}
With $A_1=-I_3$ and $A_2=sR$, the depth residual covariance is $C^d_p=\sum_w A_w\Sigma^{d,(w)}_p A_w^\top$. For each shared error, sum the Jacobians over pixels of frame $m$ \emph{before} propagating its covariance:
\[
 \begin{aligned}
 G_{w,m} &= \sum_{p\in m}U_p A_w J^{(w)}_p, &
 G^K_{w,m} &= \sum_{p\in m}U_p A_w B^{(w)}_p .
 \end{aligned}
\]
For a skip edge of VGGT-Bridge, the reference term adds $\mathbf{G}_{\mathrm{ref}}\big(\boldsymbol{\Sigma}^{\mathrm{focal}}_{\mathrm{ref},1}+\boldsymbol{\Sigma}^{\mathrm{focal}}_{\mathrm{ref},2}\big)\mathbf{G}_{\mathrm{ref}}^{\top}$ to $V_K$, with $\mathbf{G}_{\mathrm{ref}}=\sum_p U_p\mathbf{Q}_p$.
No robust kernel is applied to the point residuals.
We initialize Levenberg--Marquardt with the weighted Umeyama solution.
If the normal equations become singular or the scale collapses to zero, we retain this initialization; this occurs in 4 of the 7,158 alignments across the three GFMs.

\subsection{VGGT-Bridge Edge Covariances}
Each VGGT-Bridge edge is formed from one registration or a composition of two registrations.
For edge transformation $C$, we express its covariance in the edge residual coordinates:
\begin{center}
\footnotesize
\begin{tabular}{@{}lll@{}}
\toprule
Edge & Composition & $\boldsymbol{\Sigma}_e$ \\
\midrule
sequential & $C=\hat S$ & $\mathrm{Ad}_C\,\boldsymbol{\Sigma}_{\boldsymbol{\xi}}\,\mathrm{Ad}_C^{\top}$ \\
loop & $C=\hat S_j\hat S_i^{-1}$ & $\mathrm{Ad}_C\,\mathrm{Ad}_{\hat S_i}\big(\boldsymbol{\Sigma}_{\boldsymbol{\xi},i}+\boldsymbol{\Sigma}_{\boldsymbol{\xi},j}\big)\mathrm{Ad}_{\hat S_i}^{\top}\mathrm{Ad}_C^{\top}$ \\
skip & $C=\hat S_a^{-1}\hat S_b$ & $\mathrm{Ad}_C\,\boldsymbol{\Sigma}_{\boldsymbol{\xi},b}\,\mathrm{Ad}_C^{\top}+\boldsymbol{\Sigma}_{\boldsymbol{\xi},a}$ \\
\bottomrule
\end{tabular}
\end{center}
The two registrations forming a loop or skip edge use disjoint frames and are treated as independent.
We also assume independence between graph edges and whiten each edge residual using $\boldsymbol{\Sigma}_e^{-1/2}$.

\subsection{FoundationVIO Factor Covariances}
The pose head predicts $\boldsymbol{\Sigma}_{ij}$ for a right perturbation $\boldsymbol{\xi}=(\boldsymbol{\rho},\boldsymbol{\phi})$ of the relative camera pose $\mathbf{P}_i^{-1}\mathbf{P}_j$.
FoundationVIO instead uses the relative body pose $\mathbf{Z}=\mathbf{T}_{BC}\mathbf{P}_i^{-1}\mathbf{P}_j\mathbf{T}_{CB}$, where $\mathbf{T}_{BC}$ maps camera coordinates to body coordinates.
Writing body poses as $(R_i,t_i)$, the residual is
\begin{equation}
 r_{ij}=\begin{bmatrix}
 \operatorname{Log}(Z_R^\top R_i^\top R_j)\\
 Z_t-e^{\lambda} R_i^\top(t_j-t_i)
 \end{bmatrix}.
 \label{eq:supp_scale_pose}
\end{equation}
The window log-scale $\lambda$ maps metric translations to reconstruction units.
In the body frame, $\mathbf{Z}^{\mathrm{true}}=\mathbf{Z}\,\mathrm{Exp}(\boldsymbol{\xi}_B)$ with $\boldsymbol{\xi}_B=\mathrm{Ad}_{\mathbf{T}_{BC}}\boldsymbol{\xi}$.
At the true poses, the rotation residual is approximately $\boldsymbol{\phi}_B$ and the translation residual is approximately $-\mathbf{Z}_R\boldsymbol{\rho}_B$.
Thus $\mathbf{r}\approx\mathbf{M}_{\mathbf{Z}}\boldsymbol{\xi}_B$, with $\mathbf{M}_{\mathbf{Z}}=\left[\begin{smallmatrix}\mathbf{0}&\mathbf{I}\\-\mathbf{Z}_R&\mathbf{0}\end{smallmatrix}\right]$, and the factor covariance is
\begin{equation}
  \mathrm{Cov}(\mathbf{r})=\mathbf{M}_{\mathbf{Z}}\,\mathrm{Ad}_{\mathbf{T}_{BC}}\,\boldsymbol{\Sigma}_{ij}\,\mathrm{Ad}_{\mathbf{T}_{BC}}^{\top}\mathbf{M}_{\mathbf{Z}}^{\top}.
  \label{eq:exp_factor_cov}
\end{equation}
Sequential factors use consecutive frame pairs. Each loop factor uses a pair reconstructed from its two images, with the released Huber kernel applied after whitening.

\section{Downstream Experimental Setup}
\label{sec:supp_downstream}

All weighting schemes use the same frozen predictions within each comparison.

\subsection{Isolated Estimation Tasks}
\label{sec:supp_isolated}

\paragraph{Sim(3) alignment.}
We register consecutive windows using the same pixels in their overlapping frames. The window length and overlap are $(30,15)$ frames on EuRoC, ScaRF, and Spires, and $(60,30)$ on Aria. For the objective in \cref{sec:exp_isolated}, Identity and Heuristic use weighted Umeyama; \method{} updates the inverse residual covariance from \cref{eq:alignment_covariance} during registration.

\paragraph{Loop closure.}
The graph estimates camera poses $T_i$ and one log-scale $\lambda_b$ per reconstruction window. Sequential pose factors use consecutive frames. Loop pose factors use image-retrieval pairs reconstructed together. Both use the scale-pose residual $r_{ij}$ in \cref{eq:supp_scale_pose}. A scale factor between windows $a,b$ compares their estimated log-scale difference with the measured log ratio,
\[
 q_{ab}=\lambda_b-\lambda_a-\log\hat s_{ab}.
\]
Here $\hat s_{ab}$ maps lengths in window $a$ to lengths in window $b$. Sequential ratios are measured on the shared frame. Let $P_t$ index pose factors and $S_t$ scale factors of type $t\in\{\mathrm{seq},\mathrm{loop}\}$. The graph minimizes
\begin{equation}
 \sum_t\left[
 \sum_{(i,j)\in P_t}\rho_t\!\left(\sqrt{r_{ij}^\top C_{ij}^{-1}r_{ij}}\right)
 +\sum_{(a,b)\in S_t}\rho_t\!\left(|q_{ab}|/\sigma_t\right)
 \right],
 \label{eq:supp_loop_task}
\end{equation}
where $\rho_{\rm seq}$ is quadratic, $\rho_{\rm loop}$ is Huber, and $(\sigma_{\rm seq},\sigma_{\rm loop})=(0.01,0.1)$. Only the pose covariance $C_{ij}$ varies across comparisons. Identity uses $I$, Heuristic uses $I/w_{ij}$ (mean confidence of the pair divided by the sequence median), and \method{} uses \cref{eq:exp_factor_cov}.

\paragraph{Visual--inertial initialization.}
Following VINS-Mono~\cite{qin2018vins}, we estimate gyroscope bias, gravity direction, and metric scale by aligning IMU preintegration with GFM relative poses over frames $k=0,\dots,n$.
Let $\mathbf{R}_k$ be the IMU orientation obtained from the predicted camera rotation and camera--IMU extrinsics.
Let $\Delta\hat{\mathbf{R}}_k$ denote the preintegrated rotation from $k$ to $k+1$, with bias Jacobian $\mathbf{J}_k$.
The gyroscope bias is estimated by
\begin{equation}
  \hat{\mathbf{b}}_g=\arg\min_{\mathbf{b}_g}\sum_{k}\Big\|\mathrm{Log}\big((\Delta\hat{\mathbf{R}}_{k}\,\mathrm{Exp}(\mathbf{J}_k\mathbf{b}_g))^{\top}\mathbf{R}_k^{\top}\mathbf{R}_{k+1}\big)\Big\|^2_{\mathbf{C}^{R}_{k}{}^{-1}} .
  \label{eq:supp_vi_bias}
\end{equation}
We then solve for velocities $\mathbf{v}_k$, gravity $\mathbf{g}$ in the first camera frame, and scale $s$.
These variables enter the preintegrated position and velocity increments $\hat{\mathbf{z}}_k=(\hat{\boldsymbol{\alpha}}_k,\hat{\boldsymbol{\beta}}_k)$ through the linear model $\hat{\mathbf{z}}_k=\mathbf{H}_k\mathbf{x}+\mathbf{n}_k$, where $\mathbf{x}=(\mathbf{v}_0,\dots,\mathbf{v}_n,\mathbf{g},s)$ and $\mathbf{H}_k$ is constructed from the predicted rotations and positions as in VINS-Mono.
The estimate is
\begin{equation}
  \hat{\mathbf{x}}=\arg\min_{\mathbf{x}}\sum_k\big\|\hat{\mathbf{z}}_k-\mathbf{H}_k\mathbf{x}\big\|^2_{\mathbf{C}_k^{-1}},
  \label{eq:supp_vi_align}
\end{equation}
followed by refinement of gravity subject to its known magnitude.
The residual covariances combine IMU and visual uncertainty:
$\mathbf{C}^{R}_{k}=\boldsymbol{\Sigma}^{\mathrm{imu}}_{R,k}+\boldsymbol{\Sigma}^{\mathrm{vis}}_{R,k}$ and $\mathbf{C}_k=\boldsymbol{\Sigma}^{\mathrm{imu}}_{k}+\boldsymbol{\Sigma}^{\mathrm{vis}}_{k}$, with visual uncertainty contributing to the position rows of the second system.

Identity uses unit covariances.
Heuristic combines calibrated IMU noise with constant visual noise.
\method{} replaces the visual noise with the covariance of each consecutive frame pair, transformed to the IMU frame as in \cref{eq:exp_factor_cov}.
Its translation component is converted to metric units using the scale from the unweighted solution.

\subsection{VGGT-Bridge}
\label{sec:supp_bridge}

We use the released chunking and loop-detection configuration: 75-frame chunks with 30-frame overlap, and additional chunk sets at strides 3 and 5.
All backbones and weighting schemes share the same chunks and detected loops.
Each backbone predicts depth, confidence, poses, and intrinsics for these chunks.
Trajectory accuracy is measured by camera-position ATE RMSE after Sim(3) Umeyama alignment.

\paragraph{Graph objective.}
The released pose graph minimizes
\begin{equation}
  \sum_e w_e\,\rho\big(\|\mathbf{r}_e\|\big),\qquad
  \rho(x)=\tfrac{\delta^2}{2}\log\!\big(1+x^2/\delta^2\big),\quad \delta=2,
  \label{eq:supp_cauchy}
\end{equation}
where $w_e$ is the scalar edge weight specified in \cref{sec:exp_systems} and $\mathbf{r}_e$ is a seven-dimensional Sim(3) residual.
With \method{}, each edge has covariance $\boldsymbol{\Sigma}_e$ from \cref{sec:supp_syscov}.
For fixed edge covariances, the Gaussian NLL reduces, up to terms constant in the graph variables, to
\begin{equation}
  \tfrac12\sum_e\mathbf{r}_e^{\top}\boldsymbol{\Sigma}_e^{-1}\mathbf{r}_e .
  \label{eq:supp_gaussian}
\end{equation}
We use this covariance-weighted objective without an additional robust kernel.
Graph connectivity and the solver remain unchanged.

\subsection{FoundationVIO}
\label{sec:supp_fvio}

FoundationVIO fuses GFM relative poses with IMU preintegration in iSAM2, using 10-frame inference windows with one-frame overlap.
The released visual noise standard deviations are 0.01\,m and $4^\circ$ for sequential factors, and 0.1\,m and $10^\circ$ for loop factors.
Loop factors also use a Huber kernel.
\method{} replaces the pose covariances according to \cref{eq:exp_factor_cov}, retaining the scale factors and loop kernel.
Every compared configuration uses the IMU noise from the corresponding dataset's calibration.
We report camera-position ATE RMSE after SE(3) Umeyama alignment.
A run is marked as divergent when its translational RPE over 1\,m path segments exceeds 5\,m/m, following FoundationVIO's criterion.

\section{Additional Ablations}
\label{sec:supp_additional_ablations}

We further examine which feature layers predict uncertainty, whether pose-error correlations are useful, and whether predicted geometry complements learned features.

\paragraph{Protocol.}
All variants are trained independently for 6,000 steps using the same training schedule and frozen backbones.
Evaluation uses the calibration datasets of the main paper, with equal weight assigned to each dataset.

\begin{figure*}[!t]
  \centering
  \includegraphics[width=\textwidth]{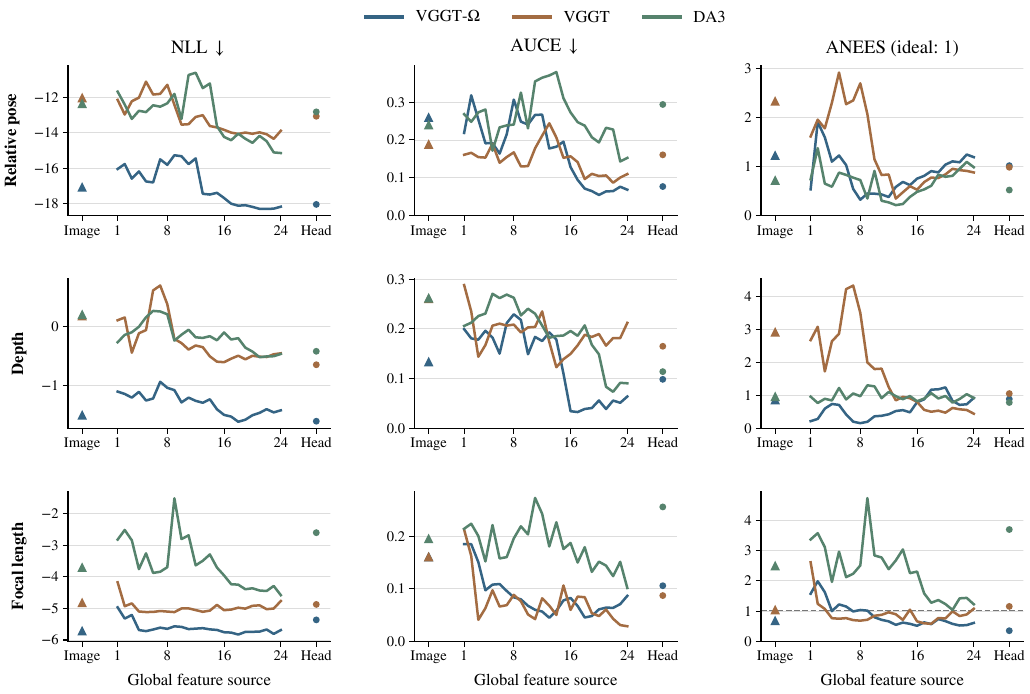}
  \caption{\textbf{Layer-wise probing of frozen global features.}
  Curves show transformer layers 1--24. Triangles mark the single-image representation (Image) and circles mark the final camera-head representation (Head).
  Image denotes the DINO image summary for VGGT-$\Omega$ and VGGT. For DA3 it denotes layer~8, the last layer that processes each image independently, so these two points coincide in value.
  We average each metric equally across the calibration datasets for its task.
  The dashed line marks ideal ANEES.}
  \label{fig:supp_layer_probing}
\end{figure*}

\subsection{Which Feature Layers Are Informative?}

We ask how well a feature from a single transformer layer predicts errors in the GFM's final reconstruction. For example, to predict pose covariance for frames $i$ and $j$, one head reads their features from layer~1 and another reads their features from layer~24. Both receive the same predicted relative pose and intrinsics, and both are trained against the same relative-pose error. We repeat this comparison at every layer. For each frame we take its camera token (the image-summary token in DA3's early layers) and, when present, the mean of that layer's register tokens.

For focal length, each head reads the chosen layer's feature for frame $i$ together with the predicted focal lengths and their window statistics. For depth, each head combines that frame feature with the same dense decoder feature at each pixel. Thus each point on a curve in \cref{fig:supp_layer_probing} represents a separately trained head using one layer. The points labeled \emph{Image} and \emph{Head} show the same comparison using single-image encoder features and the final features used to predict camera parameters, respectively.

Later transformer features generally give better calibration (\cref{fig:supp_layer_probing}). For every task and backbone, median AUCE across layers 17--24 is lower than across layers 1--8. Our pose and focal heads use the last transformer layer (layer~24). On all three backbones, this layer gives lower pose NLL and AUCE and lower focal AUCE than both the single-image feature and the final camera-head feature.

Our depth covariance head uses the final camera-head feature to provide frame context at each pixel. With the dense pixel features held fixed, this frame feature gives lower depth NLL and AUCE than the single-image feature on all three backbones. Relative to layer~24, the camera-head feature gives lower NLL on VGGT-$\Omega$ and VGGT, while layer~24 gives lower NLL on DA3.

\subsection{Does a Full Pose Covariance Improve Likelihood and Calibration?}

We compare the full covariance $\Sigma=SCS$ with a block-isotropic covariance
$\Sigma=\operatorname{diag}(\sigma_t^2I_3,\sigma_r^2I_3)$.
Both heads use the same features, predicted geometry, and two residual MLP blocks. Only the covariance output changes.
The block-isotropic variant learns separate translation and rotation scales but cannot represent directional variation within either block or correlations between them.

\begin{table}[!ht]
\centering\footnotesize
\setlength{\tabcolsep}{3pt}
\caption{\textbf{Pose covariance structure and inputs.}
Equal-dataset averages on TUM-RGBD and ETH3D.
Full and block-isotropic use identical features and predicted geometry; the remaining variants retain full covariance but remove one input source.
NLL and ANEES are median window scores.
Best per backbone is in bold; ideal ANEES is 1.}
\label{tab:supp_pose_structure_inputs}
\begin{tabular}{@{}llrrr@{}}
\toprule
Backbone & Variant & NLL$\downarrow$ & AUCE$\downarrow$ & ANEES \\
\midrule
VGGT-$\Omega$ & Full & \textbf{$-$18.308} & \textbf{0.063} & 1.127 \\
VGGT-$\Omega$ & Block-isotropic & $-$17.550 & 0.073 & \textbf{1.004} \\
VGGT-$\Omega$ & Features only & $-$18.225 & 0.090 & 1.165 \\
VGGT-$\Omega$ & Geometry only & $-$12.384 & 0.409 & 2.287 \\
\midrule
VGGT & Full & \textbf{$-$14.189} & \textbf{0.110} & \textbf{0.914} \\
VGGT & Block-isotropic & $-$13.395 & 0.112 & 0.820 \\
VGGT & Features only & $-$13.940 & 0.117 & 1.092 \\
VGGT & Geometry only & $-$10.938 & 0.269 & 2.970 \\
\midrule
DA3 & Full & \textbf{$-$15.219} & \textbf{0.146} & \textbf{1.049} \\
DA3 & Block-isotropic & $-$14.634 & 0.161 & 0.880 \\
DA3 & Features only & $-$15.199 & 0.154 & 0.870 \\
DA3 & Geometry only & $-$10.497 & 0.358 & 0.339 \\
\bottomrule
\end{tabular}
\end{table}

In \cref{tab:supp_pose_structure_inputs}, full covariance improves both median NLL and AUCE on every backbone after averaging the two datasets.
Full covariance also lowers median NLL in five of the six individual dataset--backbone comparisons.

\subsection{Does Predicted Geometry Complement the Features?}

Keeping the full covariance output fixed, we compare camera/register features alone, predicted pair geometry alone, and their combination.
Geometry includes relative pose, intrinsics, and motion magnitudes, as in \cref{sec:supp_method}.

Combining features and geometry lowers AUCE and median NLL relative to either input alone for all three backbones (\cref{tab:supp_pose_structure_inputs}).
The gain over geometry alone is large; the gain over features alone is smaller, indicating that the frozen representation already contains much of the predictive information.

\section{Additional Results}
\label{sec:supp_additional}
\subsection{Effect of Inference Window Size}
\label{sec:supp_windows}
\begin{table*}[t]
\centering
\footnotesize
\setlength{\tabcolsep}{3.4pt}
\caption{\textbf{Sim(3) registration with different inference window sizes.} The first column gives window size/overlap in frames. Results use VGGT-$\Omega$; RRE is in degrees, RTE in centimeters, and RSE in percent. Metrics are pooled over all sequential registrations in each dataset. Best per configuration in \textbf{bold}.}
\label{tab:supp_windows}
\begin{tabular}{@{}llcccccccccccc@{}}
\toprule
 & & \multicolumn{3}{c}{EuRoC} & \multicolumn{3}{c}{Aria} & \multicolumn{3}{c}{ScaRF} & \multicolumn{3}{c}{Spires} \\
\cmidrule(lr){3-5}\cmidrule(lr){6-8}\cmidrule(lr){9-11}\cmidrule(lr){12-14}\addlinespace[0.3ex]
Window & Weighting & \metric{RRE} & \metric{RTE} & \metric{RSE} & \metric{RRE} & \metric{RTE} & \metric{RSE} & \metric{RRE} & \metric{RTE} & \metric{RSE} & \metric{RRE} & \metric{RTE} & \metric{RSE} \\
\midrule
 & Identity & \cellcolor{rk100}0.87 & \cellcolor{rk100}11.5 & \cellcolor{rk100}9.73 & \cellcolor{rk100}0.72 & \cellcolor{rk100}4.3 & \cellcolor{rk100}3.67 & \cellcolor{rk100}0.98 & \cellcolor{rk100}9.3 & \cellcolor{rk100}6.89 & \cellcolor{rk100}0.51 & \cellcolor{rk100}19.1 & \cellcolor{rk100}4.30 \\
20/10 & Heuristic & \cellcolor{rk050}0.78 & \cellcolor{rk050}8.7 & \cellcolor{rk050}9.29 & \cellcolor{rk050}0.62 & \cellcolor{rk050}3.0 & \cellcolor{rk050}2.92 & \cellcolor{rk050}0.91 & \cellcolor{rk050}7.5 & \cellcolor{rk050}6.34 & \cellcolor{rk050}0.43 & \cellcolor{rk050}10.3 & \cellcolor{rk050}3.23 \\
\rowcolor{oursrow} & \method{} & \cellcolor{rk000}\best{0.48} & \cellcolor{rk000}\best{5.3} & \cellcolor{rk000}\best{6.53} & \cellcolor{rk000}\best{0.44} & \cellcolor{rk000}\best{1.9} & \cellcolor{rk000}\best{2.53} & \cellcolor{rk000}\best{0.70} & \cellcolor{rk000}\best{5.9} & \cellcolor{rk000}\best{6.13} & \cellcolor{rk000}\best{0.27} & \cellcolor{rk000}\best{3.5} & \cellcolor{rk000}\best{2.02} \\
\cmidrule(l){2-14}
 & Identity & \cellcolor{rk100}0.95 & \cellcolor{rk100}12.2 & \cellcolor{rk100}9.27 & \cellcolor{rk100}0.79 & \cellcolor{rk100}4.9 & \cellcolor{rk100}3.46 & \cellcolor{rk100}1.02 & \cellcolor{rk100}10.7 & \cellcolor{rk100}3.87 & \cellcolor{rk100}0.51 & \cellcolor{rk100}22.1 & \cellcolor{rk100}4.33 \\
30/15 & Heuristic & \cellcolor{rk050}0.86 & \cellcolor{rk050}9.5 & \cellcolor{rk050}8.77 & \cellcolor{rk050}0.70 & \cellcolor{rk050}3.3 & \cellcolor{rk050}2.54 & \cellcolor{rk050}0.99 & \cellcolor{rk050}9.6 & \cellcolor{rk050}3.41 & \cellcolor{rk050}0.45 & \cellcolor{rk050}12.4 & \cellcolor{rk050}3.09 \\
\rowcolor{oursrow} & \method{} & \cellcolor{rk000}\best{0.61} & \cellcolor{rk000}\best{6.6} & \cellcolor{rk000}\best{6.18} & \cellcolor{rk000}\best{0.55} & \cellcolor{rk000}\best{2.4} & \cellcolor{rk000}\best{2.20} & \cellcolor{rk000}\best{0.85} & \cellcolor{rk000}\best{8.2} & \cellcolor{rk000}\best{3.31} & \cellcolor{rk000}\best{0.32} & \cellcolor{rk000}\best{4.8} & \cellcolor{rk000}\best{1.80} \\
\cmidrule(l){2-14}
 & Identity & \cellcolor{rk100}1.11 & \cellcolor{rk100}12.9 & \cellcolor{rk100}3.94 & \cellcolor{rk100}0.94 & \cellcolor{rk100}6.2 & \cellcolor{rk100}2.80 & \cellcolor{rk100}2.30 & \cellcolor{rk100}18.6 & \cellcolor{rk100}4.06 & \cellcolor{rk100}0.64 & \cellcolor{rk100}27.0 & \cellcolor{rk100}3.81 \\
60/30 & Heuristic & \cellcolor{rk050}1.07 & \cellcolor{rk050}10.7 & \cellcolor{rk050}3.45 & \cellcolor{rk050}0.88 & \cellcolor{rk050}4.9 & \cellcolor{rk050}2.13 & \cellcolor{rk050}2.29 & \cellcolor{rk050}17.1 & \cellcolor{rk000}\best{3.74} & \cellcolor{rk050}0.60 & \cellcolor{rk050}16.2 & \cellcolor{rk050}2.66 \\
\rowcolor{oursrow} & \method{} & \cellcolor{rk000}\best{0.89} & \cellcolor{rk000}\best{9.1} & \cellcolor{rk000}\best{2.22} & \cellcolor{rk000}\best{0.79} & \cellcolor{rk000}\best{4.2} & \cellcolor{rk000}\best{1.96} & \cellcolor{rk000}\best{1.99} & \cellcolor{rk000}\best{15.0} & \cellcolor{rk050}3.95 & \cellcolor{rk000}\best{0.51} & \cellcolor{rk000}\best{9.6} & \cellcolor{rk000}\best{1.77} \\
\cmidrule(l){2-14}
 & Identity & \cellcolor{rk100}1.34 & \cellcolor{rk100}14.7 & \cellcolor{rk100}4.69 & \cellcolor{rk100}1.14 & \cellcolor{rk100}6.9 & \cellcolor{rk100}2.80 & \cellcolor{rk050}5.88 & \cellcolor{rk100}36.6 & \cellcolor{rk100}7.92 & \cellcolor{rk100}0.78 & \cellcolor{rk100}33.4 & \cellcolor{rk100}5.01 \\
75/30 & Heuristic & \cellcolor{rk050}1.30 & \cellcolor{rk050}12.7 & \cellcolor{rk050}4.28 & \cellcolor{rk050}1.08 & \cellcolor{rk050}5.8 & \cellcolor{rk050}2.21 & \cellcolor{rk100}5.92 & \cellcolor{rk050}35.4 & \cellcolor{rk050}7.43 & \cellcolor{rk050}0.74 & \cellcolor{rk050}21.0 & \cellcolor{rk050}3.57 \\
\rowcolor{oursrow} & \method{} & \cellcolor{rk000}\best{1.07} & \cellcolor{rk000}\best{10.5} & \cellcolor{rk000}\best{2.76} & \cellcolor{rk000}\best{1.00} & \cellcolor{rk000}\best{5.1} & \cellcolor{rk000}\best{2.06} & \cellcolor{rk000}\best{5.73} & \cellcolor{rk000}\best{32.2} & \cellcolor{rk000}\best{7.39} & \cellcolor{rk000}\best{0.66} & \cellcolor{rk000}\best{12.4} & \cellcolor{rk000}\best{2.36} \\
\bottomrule
\end{tabular}

\end{table*}

We repeat the VGGT-$\Omega$ registration experiment with 20-, 30-, 60-, and 75-frame windows, using overlaps of 10, 15, 30, and 30 frames, respectively.
All other settings follow \cref{sec:supp_isolated}.
As shown in \cref{tab:supp_windows}, rotation and translation errors generally increase with window size.
\method{} achieves the lowest rotation and translation errors in every dataset--window configuration and the lowest scale error in 15 of 16 configurations.

\paragraph{Long windows on ScaRF.}
The scale-error exception occurs on ScaRF with 60-frame windows.
Inspection reveals inconsistent geometry around doorway transitions: within one 60-frame window and two 75-frame windows, frames after the doorway are rotated relative to earlier frames by up to $40^\circ$ and $130$--$170^\circ$, respectively.
The next window reconstructs the same frames correctly, producing large disagreement in the overlap.
These registrations dominate the mean errors for all three weighting schemes.
\method{} has slightly higher scale error than Heuristic with 60 frames (3.95 \vs 3.74) and slightly lower error with 75 frames (7.39 \vs 7.43).
No registration exceeds $30^\circ$ rotation error with 20- or 30-frame windows.

\paragraph{Short windows and static cameras.}
The reference Sim(3) depends on a scale fitted between predicted and ground-truth camera positions in each window.
This scale is poorly constrained when camera motion is small, as at the beginning of the EuRoC Machine Hall sequences.
Such windows contribute large scale errors under every weighting, particularly for EuRoC with 20 or 30 frames and ScaRF with 20 frames.
As a diagnostic, excluding registrations for which either window's ground-truth camera positions all lie within 0.3\,m of their centroid reduces the EuRoC 30-frame scale errors to 2.64, 2.28, and 1.75, and the ScaRF 20-frame errors to 3.81, 3.24, and 3.03 (Identity, Heuristic, and \method{}, respectively).
The reported tables retain all registrations.

\subsection{Reconstruction on Driving Sequences}
\label{sec:supp_qualitative}

We additionally evaluate VGGT-Bridge with VGGT-$\Omega$ and \method{} on KITTI odometry~\cite{geiger2012kitti} and 4Seasons~\cite{wenzel20204seasons}, neither of which is used to train the covariance heads.
The system settings follow \cref{sec:supp_bridge}.
For KITTI, we use the left color camera at 10\,Hz; for 4Seasons, we use the left undistorted camera at the reference-pose timestamps, approximately 10\,Hz.
We evaluate the complete KITTI sequence 06 (1,101 frames), frames 700--1099 of KITTI sequence 08, and 400-frame segments of the 4Seasons Office Loop and Neighborhood recordings.

\begin{figure}[!ht]
  \centering
  \includegraphics[width=\linewidth]{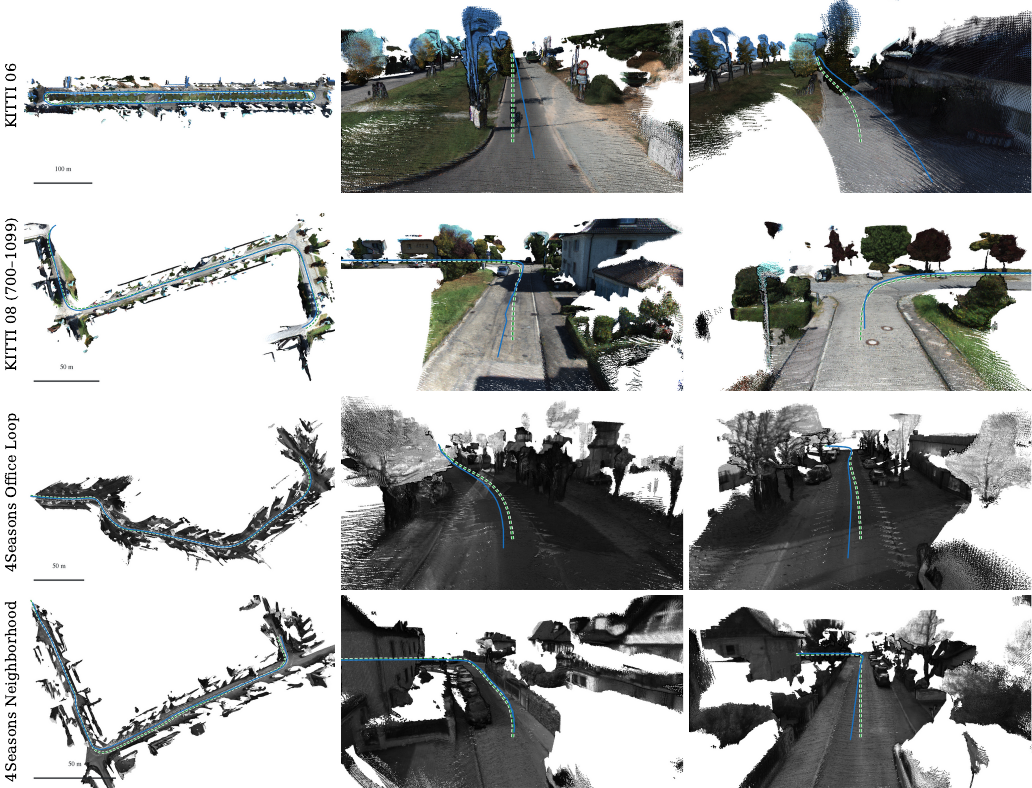}
  \caption{\textbf{VGGT-Bridge with \method{} on driving sequences.}
  Each row shows a top-down map (left) and two views from behind and above the estimated camera (middle and right).
  Blue denotes the estimated trajectory; dashed green denotes ground truth.
  Maps are formed by back-projecting the predicted depth and applying the optimized chunk poses.}
  \label{fig:supp_driving}
\end{figure}

\Cref{fig:supp_driving} shows the reconstructed maps and trajectories.
The ATE is 2.09\,m on KITTI 06 (1.2\,km, with loop closure), 1.11\,m on KITTI 08 (315\,m), 0.94\,m on Office Loop (346\,m), and 1.66\,m on Neighborhood (246\,m).

\section{Per-Sequence Results}
\label{sec:supp_perseq}

The following tables give the per-sequence results underlying \cref{tab:isolated,tab:systems}.
They are grouped by task: Sim(3) registration, loop closure, visual--inertial initialization, and SLAM systems.
\Cref{sec:supp_downstream} defines the objectives and weighting schemes.
Within each backbone, all methods use the same geometric predictions.
The loop-closure tables also include the common open-loop result in gray.

The \emph{Mean} column uses the same aggregation as the main paper.
For registration, it pools all edges, weighting each sequence in proportion to its number of edges.
For trajectory metrics, it is the arithmetic mean of the per-sequence scores.

\paragraph{Sequence names.}
In EuRoC, MH01--MH05 denote the Machine Hall sequences and V101--V203 the Vicon Room sequences.
In Spires, CC02, CC03, and CC05 denote Christ Church sequences 02, 03, and 05.

\onecolumn
\makeatletter\global\@colroom\@colht\global\vsize\@colht\makeatother%
\begingroup
\makeatletter
\renewenvironment{table*}[1][]{\def\@captype{table}}{}
\let\footnotesize\scriptsize%
\renewcommand{\arraystretch}{0.88}
\newcommand{\suppfull}[1]{\par\addvspace{\medskipamount}\noindent\begin{minipage}{\textwidth}#1\end{minipage}\par\addvspace{\medskipamount}}
\newcommand{\supppair}[2]{\par\addvspace{\medskipamount}\noindent\begin{minipage}[t]{.495\textwidth}#1\end{minipage}\hfill\begin{minipage}[t]{.495\textwidth}#2\end{minipage}\par\addvspace{\medskipamount}}
\makeatother
\raggedbottom
\suppfull{\begin{table*}[t]
\centering
\footnotesize
\setlength{\tabcolsep}{2.6pt}
\caption{\textbf{Per-sequence Sim(3) registration on EuRoC.} Settings follow \cref{sec:supp_downstream}; aggregation follows \cref{sec:supp_perseq}.}
\label{tab:supp_euroc_align}
\begin{tabular}{@{}llcccccccccccc@{}}
\toprule
Backbone & Weighting & MH01 & MH02 & MH03 & MH04 & MH05 & V101 & V102 & V103 & V201 & V202 & V203 & Mean \\
\midrule
\multicolumn{14}{@{}l}{\textit{RRE ($^\circ$)}} \\
\cmidrule(r){1-2}
 & Identity & \cellcolor{rk100}0.68 & \cellcolor{rk100}0.69 & \cellcolor{rk100}1.00 & \cellcolor{rk100}1.04 & \cellcolor{rk100}0.94 & \cellcolor{rk100}1.35 & \cellcolor{rk100}0.93 & \cellcolor{rk100}0.86 & \cellcolor{rk100}0.74 & \cellcolor{rk050}0.92 & \cellcolor{rk100}1.54 & \cellcolor{rk100}0.95 \\
VGGT-$\Omega$ & Heuristic & \cellcolor{rk050}0.61 & \cellcolor{rk050}0.57 & \cellcolor{rk050}0.82 & \cellcolor{rk050}0.80 & \cellcolor{rk050}0.67 & \cellcolor{rk050}1.34 & \cellcolor{rk050}0.89 & \cellcolor{rk050}0.83 & \cellcolor{rk050}0.72 & \cellcolor{rk050}0.92 & \cellcolor{rk050}1.53 & \cellcolor{rk050}0.86 \\
\rowcolor{oursrow} & \method{} & \cellcolor{rk000}\best{0.29} & \cellcolor{rk000}\best{0.29} & \cellcolor{rk000}\best{0.44} & \cellcolor{rk000}\best{0.41} & \cellcolor{rk000}\best{0.41} & \cellcolor{rk000}\best{1.12} & \cellcolor{rk000}\best{0.78} & \cellcolor{rk000}\best{0.72} & \cellcolor{rk000}\best{0.48} & \cellcolor{rk000}\best{0.87} & \cellcolor{rk000}\best{1.28} & \cellcolor{rk000}\best{0.61} \\
\cmidrule(l){2-14}
 & Identity & \cellcolor{rk100}2.29 & \cellcolor{rk100}1.80 & \cellcolor{rk100}2.58 & \cellcolor{rk100}3.31 & \cellcolor{rk100}2.43 & \cellcolor{rk100}2.65 & \cellcolor{rk100}2.53 & \cellcolor{rk100}3.39 & \cellcolor{rk100}2.19 & \cellcolor{rk100}2.37 & \cellcolor{rk050}4.13 & \cellcolor{rk100}2.62 \\
VGGT & Heuristic & \cellcolor{rk050}2.18 & \cellcolor{rk050}1.70 & \cellcolor{rk050}2.57 & \cellcolor{rk050}3.15 & \cellcolor{rk050}2.40 & \cellcolor{rk050}2.61 & \cellcolor{rk050}2.52 & \cellcolor{rk050}3.33 & \cellcolor{rk050}2.13 & \cellcolor{rk050}2.34 & \cellcolor{rk100}4.17 & \cellcolor{rk050}2.56 \\
\bottomrule
\end{tabular}

\end{table*}
}
\suppfull{\begin{table*}[t]
\centering
\footnotesize
\setlength{\tabcolsep}{3.1pt}
\caption{\textbf{Per-sequence Sim(3) registration on Aria.} Settings follow \cref{sec:supp_downstream}; aggregation follows \cref{sec:supp_perseq}.}
\label{tab:supp_aria_align}
\begin{tabular}{@{}llcccccccccc@{}}
\toprule
Backbone & Weighting & AprilTag1 & AprilTag2 & BigRoom2 & BigRoom3 & BigRoom4 & NewTags1 & NewTags2 & NewTags3 & NewTags4 & Mean \\
\midrule
\multicolumn{12}{@{}l}{\textit{RRE ($^\circ$)}} \\
\cmidrule(r){1-2}
 & Identity & \cellcolor{rk100}0.84 & \cellcolor{rk100}0.74 & \cellcolor{rk100}0.69 & \cellcolor{rk100}0.67 & \cellcolor{rk100}0.76 & \cellcolor{rk100}0.80 & \cellcolor{rk050}1.48 & \cellcolor{rk100}1.12 & \cellcolor{rk100}1.25 & \cellcolor{rk100}0.94 \\
VGGT-$\Omega$ & Heuristic & \cellcolor{rk050}0.74 & \cellcolor{rk050}0.71 & \cellcolor{rk050}0.58 & \cellcolor{rk050}0.57 & \cellcolor{rk050}0.58 & \cellcolor{rk050}0.78 & \cellcolor{rk100}1.53 & \cellcolor{rk050}1.11 & \cellcolor{rk050}1.16 & \cellcolor{rk050}0.88 \\
\rowcolor{oursrow} & \method{} & \cellcolor{rk000}\best{0.68} & \cellcolor{rk000}\best{0.58} & \cellcolor{rk000}\best{0.50} & \cellcolor{rk000}\best{0.52} & \cellcolor{rk000}\best{0.53} & \cellcolor{rk000}\best{0.74} & \cellcolor{rk000}\best{1.36} & \cellcolor{rk000}\best{1.00} & \cellcolor{rk000}\best{1.07} & \cellcolor{rk000}\best{0.79} \\
\cmidrule(l){2-12}
 & Identity & \cellcolor{rk100}8.70 & \cellcolor{rk100}6.34 & \cellcolor{rk050}2.65 & \cellcolor{rk100}3.95 & \cellcolor{rk100}2.62 & \cellcolor{rk100}4.64 & \cellcolor{rk050}16.3 & \cellcolor{rk100}14.1 & \cellcolor{rk050}17.1 & \cellcolor{rk100}9.35 \\
VGGT & Heuristic & \cellcolor{rk050}8.67 & \cellcolor{rk050}6.30 & \cellcolor{rk050}2.65 & \cellcolor{rk050}3.90 & \cellcolor{rk050}2.56 & \cellcolor{rk050}4.63 & \cellcolor{rk100}16.4 & \cellcolor{rk050}13.9 & \cellcolor{rk050}17.1 & \cellcolor{rk050}9.32 \\
\rowcolor{oursrow} & \method{} & \cellcolor{rk000}\best{7.44} & \cellcolor{rk000}\best{4.73} & \cellcolor{rk000}\best{1.81} & \cellcolor{rk000}\best{2.43} & \cellcolor{rk000}\best{1.90} & \cellcolor{rk000}\best{3.99} & \cellcolor{rk000}\best{15.6} & \cellcolor{rk000}\best{12.5} & \cellcolor{rk000}\best{14.0} & \cellcolor{rk000}\best{7.84} \\
\bottomrule
\end{tabular}

\end{table*}
}
\supppair{\begin{table*}[t]
\centering
\footnotesize
\setlength{\tabcolsep}{3.2pt}
\caption{\textbf{Per-sequence Sim(3) registration on ScaRF.} Settings follow \cref{sec:supp_downstream}; aggregation follows \cref{sec:supp_perseq}.}
\label{tab:supp_scarf_align}
\begin{tabular}{@{}llcccccc@{}}
\toprule
Backbone & Weighting & r01 & r02 & r03 & r04 & r05 & Mean \\
\midrule
\multicolumn{8}{@{}l}{\textit{RRE ($^\circ$)}} \\
\cmidrule(r){1-2}
 & Identity & \cellcolor{rk100}1.15 & \cellcolor{rk050}1.29 & \cellcolor{rk050}1.03 & \cellcolor{rk100}0.96 & \cellcolor{rk100}0.63 & \cellcolor{rk100}1.02 \\
VGGT-$\Omega$ & Heuristic & \cellcolor{rk050}1.10 & \cellcolor{rk100}1.30 & \cellcolor{rk100}1.05 & \cellcolor{rk050}0.87 & \cellcolor{rk050}0.57 & \cellcolor{rk050}0.99 \\
\rowcolor{oursrow} & \method{} & \cellcolor{rk000}\best{0.84} & \cellcolor{rk000}\best{1.20} & \cellcolor{rk000}\best{0.91} & \cellcolor{rk000}\best{0.75} & \cellcolor{rk000}\best{0.50} & \cellcolor{rk000}\best{0.85} \\
\cmidrule(l){2-8}
 & Identity & \cellcolor{rk100}4.38 & \cellcolor{rk100}25.6 & \cellcolor{rk050}12.3 & \cellcolor{rk100}4.62 & \cellcolor{rk100}2.36 & \cellcolor{rk100}9.66 \\
VGGT & Heuristic & \cellcolor{rk050}4.36 & \cellcolor{rk050}25.4 & \cellcolor{rk050}12.3 & \cellcolor{rk050}4.60 & \cellcolor{rk050}2.30 & \cellcolor{rk050}9.62 \\
\rowcolor{oursrow} & \method{} & \cellcolor{rk000}\best{2.35} & \cellcolor{rk000}\best{19.9} & \cellcolor{rk000}\best{9.17} & \cellcolor{rk000}\best{3.29} & \cellcolor{rk000}\best{1.53} & \cellcolor{rk000}\best{7.10} \\
\bottomrule
\end{tabular}

\end{table*}
}{\begin{table*}[t]
\centering
\footnotesize
\setlength{\tabcolsep}{6pt}
\caption{\textbf{Per-sequence Sim(3) registration on Spires.} Settings follow \cref{sec:supp_downstream}; aggregation follows \cref{sec:supp_perseq}.}
\label{tab:supp_spires_align}
\begin{tabular}{@{}llcccc@{}}
\toprule
Backbone & Weighting & CC02 & CC03 & CC05 & Mean \\
\midrule
\multicolumn{6}{@{}l}{\textit{RRE ($^\circ$)}} \\
\cmidrule(r){1-2}
 & Identity & \cellcolor{rk100}0.49 & \cellcolor{rk100}0.52 & \cellcolor{rk100}0.53 & \cellcolor{rk100}0.51 \\
VGGT-$\Omega$ & Heuristic & \cellcolor{rk050}0.43 & \cellcolor{rk050}0.49 & \cellcolor{rk050}0.45 & \cellcolor{rk050}0.45 \\
\rowcolor{oursrow} & \method{} & \cellcolor{rk000}\best{0.29} & \cellcolor{rk000}\best{0.34} & \cellcolor{rk000}\best{0.33} & \cellcolor{rk000}\best{0.32} \\
\cmidrule(l){2-6}
 & Identity & \cellcolor{rk100}1.93 & \cellcolor{rk050}2.22 & \cellcolor{rk100}1.78 & \cellcolor{rk100}1.91 \\
VGGT & Heuristic & \cellcolor{rk050}1.89 & \cellcolor{rk100}2.24 & \cellcolor{rk050}1.76 & \cellcolor{rk050}1.89 \\
\rowcolor{oursrow} & \method{} & \cellcolor{rk000}\best{1.38} & \cellcolor{rk000}\best{1.39} & \cellcolor{rk000}\best{1.28} & \cellcolor{rk000}\best{1.33} \\
\bottomrule
\end{tabular}

\end{table*}
}
\suppfull{\begin{table*}[t]
\centering
\footnotesize
\setlength{\tabcolsep}{2.6pt}
\caption{\textbf{Per-sequence loop closure on EuRoC.} Settings follow \cref{sec:supp_downstream}; aggregation follows \cref{sec:supp_perseq}.}
\label{tab:supp_euroc_loop}
\begin{tabular}{@{}llcccccccccccc@{}}
\toprule
Backbone & Weighting & MH01 & MH02 & MH03 & MH04 & MH05 & V101 & V102 & V103 & V201 & V202 & V203 & Mean \\
\midrule
\multicolumn{14}{@{}l}{\textit{ATE (m)}} \\
\cmidrule(r){1-2}\addlinespace[0.3ex]
 & \textcolor{black!70}{\textit{no loop closure}} & \textcolor{black!70}{0.450} & \textcolor{black!70}{0.279} & \textcolor{black!70}{0.555} & \textcolor{black!70}{0.500} & \textcolor{black!70}{0.619} & \textcolor{black!70}{0.132} & \textcolor{black!70}{0.241} & \textcolor{black!70}{0.238} & \textcolor{black!70}{0.207} & \textcolor{black!70}{0.216} & \textcolor{black!70}{0.392} & \textcolor{black!70}{0.348} \\
 & Identity & \cellcolor{rk100}0.351 & \cellcolor{rk100}0.410 & \cellcolor{rk100}0.357 & \cellcolor{rk100}0.363 & \cellcolor{rk050}0.442 & \cellcolor{rk050}0.238 & \cellcolor{rk050}0.169 & \cellcolor{rk100}0.194 & \cellcolor{rk050}0.386 & \cellcolor{rk050}0.288 & \cellcolor{rk050}0.265 & \cellcolor{rk050}0.315 \\
VGGT-$\Omega$ & Heuristic & \cellcolor{rk050}0.349 & \cellcolor{rk050}0.403 & \cellcolor{rk050}0.348 & \cellcolor{rk050}0.359 & \cellcolor{rk100}0.492 & \cellcolor{rk100}0.240 & \cellcolor{rk050}0.169 & \cellcolor{rk050}0.193 & \cellcolor{rk100}0.390 & \cellcolor{rk100}0.300 & \cellcolor{rk100}0.268 & \cellcolor{rk100}0.319 \\
\rowcolor{oursrow} & \method{} & \cellcolor{rk000}\best{0.163} & \cellcolor{rk000}\best{0.116} & \cellcolor{rk000}\best{0.267} & \cellcolor{rk000}\best{0.357} & \cellcolor{rk000}\best{0.321} & \cellcolor{rk000}\best{0.071} & \cellcolor{rk000}\best{0.139} & \cellcolor{rk000}\best{0.172} & \cellcolor{rk000}\best{0.094} & \cellcolor{rk000}\best{0.127} & \cellcolor{rk000}\best{0.144} & \cellcolor{rk000}\best{0.179} \\
\bottomrule
\end{tabular}

\end{table*}
}
\suppfull{\begin{table*}[t]
\centering
\footnotesize
\setlength{\tabcolsep}{3.1pt}
\caption{\textbf{Per-sequence loop closure on Aria.} Settings follow \cref{sec:supp_downstream}; aggregation follows \cref{sec:supp_perseq}.}
\label{tab:supp_aria_loop}
\begin{tabular}{@{}llcccccccccc@{}}
\toprule
Backbone & Weighting & AprilTag1 & AprilTag2 & BigRoom2 & BigRoom3 & BigRoom4 & NewTags1 & NewTags2 & NewTags3 & NewTags4 & Mean \\
\midrule
\multicolumn{12}{@{}l}{\textit{ATE (m)}} \\
\cmidrule(r){1-2}\addlinespace[0.3ex]
 & \textcolor{black!70}{\textit{no loop closure}} & \textcolor{black!70}{1.078} & \textcolor{black!70}{1.274} & \textcolor{black!70}{0.291} & \textcolor{black!70}{0.316} & \textcolor{black!70}{0.256} & \textcolor{black!70}{0.853} & \textcolor{black!70}{1.101} & \textcolor{black!70}{1.910} & \textcolor{black!70}{1.275} & \textcolor{black!70}{0.928} \\
 & Identity & \cellcolor{rk100}0.875 & \cellcolor{rk100}4.380 & \cellcolor{rk100}0.285 & \cellcolor{rk050}0.390 & \cellcolor{rk050}0.457 & \cellcolor{rk100}0.972 & \cellcolor{rk050}1.155 & \cellcolor{rk100}6.864 & \cellcolor{rk050}3.155 & \cellcolor{rk100}2.059 \\
VGGT-$\Omega$ & Heuristic & \cellcolor{rk050}0.862 & \cellcolor{rk050}4.360 & \cellcolor{rk050}0.284 & \cellcolor{rk050}0.390 & \cellcolor{rk050}0.457 & \cellcolor{rk050}0.967 & \cellcolor{rk100}1.173 & \cellcolor{rk050}6.619 & \cellcolor{rk100}3.240 & \cellcolor{rk050}2.039 \\
\rowcolor{oursrow} & \method{} & \cellcolor{rk000}\best{0.207} & \cellcolor{rk000}\best{0.173} & \cellcolor{rk000}\best{0.153} & \cellcolor{rk000}\best{0.130} & \cellcolor{rk000}\best{0.131} & \cellcolor{rk000}\best{0.777} & \cellcolor{rk000}\best{0.255} & \cellcolor{rk000}\best{0.435} & \cellcolor{rk000}\best{0.844} & \cellcolor{rk000}\best{0.345} \\
\cmidrule(l){2-12}\addlinespace[0.3ex]
 & \textcolor{black!70}{\textit{no loop closure}} & \textcolor{black!70}{6.138} & \textcolor{black!70}{2.419} & \textcolor{black!70}{0.608} & \textcolor{black!70}{0.610} & \textcolor{black!70}{0.627} & \textcolor{black!70}{2.454} & \textcolor{black!70}{5.125} & \textcolor{black!70}{2.757} & \textcolor{black!70}{5.586} & \textcolor{black!70}{2.925} \\
\bottomrule
\end{tabular}

\end{table*}
}
\supppair{\begin{table*}[t]
\centering
\footnotesize
\setlength{\tabcolsep}{3.2pt}
\caption{\textbf{Per-sequence loop closure on ScaRF.} Settings follow \cref{sec:supp_downstream}; aggregation follows \cref{sec:supp_perseq}.}
\label{tab:supp_scarf_loop}
\begin{tabular}{@{}llcccccc@{}}
\toprule
Backbone & Weighting & r01 & r02 & r03 & r04 & r05 & Mean \\
\midrule
\multicolumn{8}{@{}l}{\textit{ATE (m)}} \\
\cmidrule(r){1-2}\addlinespace[0.3ex]
 & \textcolor{black!70}{\textit{no loop closure}} & \textcolor{black!70}{1.547} & \textcolor{black!70}{1.562} & \textcolor{black!70}{0.885} & \textcolor{black!70}{0.591} & \textcolor{black!70}{0.520} & \textcolor{black!70}{1.021} \\
 & Identity & \cellcolor{rk050}0.534 & \cellcolor{rk100}0.234 & \cellcolor{rk050}0.254 & \cellcolor{rk100}1.378 & \cellcolor{rk050}0.198 & \cellcolor{rk050}0.520 \\
VGGT-$\Omega$ & Heuristic & \cellcolor{rk100}0.554 & \cellcolor{rk050}0.227 & \cellcolor{rk100}0.259 & \cellcolor{rk050}1.362 & \cellcolor{rk100}0.203 & \cellcolor{rk100}0.521 \\
\rowcolor{oursrow} & \method{} & \cellcolor{rk000}\best{0.251} & \cellcolor{rk000}\best{0.125} & \cellcolor{rk000}\best{0.125} & \cellcolor{rk000}\best{0.169} & \cellcolor{rk000}\best{0.097} & \cellcolor{rk000}\best{0.153} \\
\cmidrule(l){2-8}\addlinespace[0.3ex]
 & \textcolor{black!70}{\textit{no loop closure}} & \textcolor{black!70}{4.358} & \textcolor{black!70}{1.910} & \textcolor{black!70}{1.928} & \textcolor{black!70}{2.301} & \textcolor{black!70}{1.282} & \textcolor{black!70}{2.356} \\
 & Identity & \cellcolor{rk050}1.322 & \cellcolor{rk000}\best{0.332} & \cellcolor{rk050}0.379 & \cellcolor{rk100}4.328 & \cellcolor{rk050}0.253 & \cellcolor{rk100}1.323 \\
\bottomrule
\end{tabular}

\end{table*}
}{\begin{table*}[t]
\centering
\footnotesize
\setlength{\tabcolsep}{6pt}
\caption{\textbf{Per-sequence loop closure on Spires.} Settings follow \cref{sec:supp_downstream}; aggregation follows \cref{sec:supp_perseq}.}
\label{tab:supp_spires_loop}
\begin{tabular}{@{}llcccc@{}}
\toprule
Backbone & Weighting & CC02 & CC03 & CC05 & Mean \\
\midrule
\multicolumn{6}{@{}l}{\textit{ATE (m)}} \\
\cmidrule(r){1-2}\addlinespace[0.3ex]
 & \textcolor{black!70}{\textit{no loop closure}} & \textcolor{black!70}{9.953} & \textcolor{black!70}{1.849} & \textcolor{black!70}{21.67} & \textcolor{black!70}{11.16} \\
 & Identity & \cellcolor{rk050}23.01 & \cellcolor{rk100}1.661 & \cellcolor{rk050}20.36 & \cellcolor{rk050}15.01 \\
VGGT-$\Omega$ & Heuristic & \cellcolor{rk100}24.45 & \cellcolor{rk050}1.645 & \cellcolor{rk100}30.58 & \cellcolor{rk100}18.89 \\
\rowcolor{oursrow} & \method{} & \cellcolor{rk000}\best{0.788} & \cellcolor{rk000}\best{0.473} & \cellcolor{rk000}\best{4.297} & \cellcolor{rk000}\best{1.852} \\
\cmidrule(l){2-6}\addlinespace[0.3ex]
 & \textcolor{black!70}{\textit{no loop closure}} & \textcolor{black!70}{33.08} & \textcolor{black!70}{16.73} & \textcolor{black!70}{36.74} & \textcolor{black!70}{28.85} \\
 & Identity & \cellcolor{rk050}18.99 & \cellcolor{rk100}2.012 & \cellcolor{rk050}27.55 & \cellcolor{rk050}16.19 \\
\bottomrule
\end{tabular}

\end{table*}
}
\suppfull{\begin{table*}[t]
\centering
\footnotesize
\setlength{\tabcolsep}{2.6pt}
\caption{\textbf{Per-sequence visual--inertial initialization on EuRoC.} Settings follow \cref{sec:supp_downstream}; aggregation follows \cref{sec:supp_perseq}.}
\label{tab:supp_euroc_vi}
\begin{tabular}{@{}llcccccccccccc@{}}
\toprule
Backbone & Weighting & MH01 & MH02 & MH03 & MH04 & MH05 & V101 & V102 & V103 & V201 & V202 & V203 & Mean \\
\midrule
\multicolumn{14}{@{}l}{\textit{$e_s$ (\%)}} \\
\cmidrule(r){1-2}
 & Identity & \cellcolor{rk100}33.0 & \cellcolor{rk100}36.2 & \cellcolor{rk000}\best{13.3} & \cellcolor{rk100}51.3 & \cellcolor{rk050}26.3 & \cellcolor{rk100}8.93 & \cellcolor{rk100}7.56 & \cellcolor{rk100}9.24 & \cellcolor{rk050}9.22 & \cellcolor{rk000}\best{5.00} & \cellcolor{rk050}5.20 & \cellcolor{rk100}18.6 \\
VGGT-$\Omega$ & Heuristic & \cellcolor{rk050}23.2 & \cellcolor{rk050}26.6 & \cellcolor{rk050}14.7 & \cellcolor{rk050}38.9 & \cellcolor{rk100}26.8 & \cellcolor{rk000}\best{7.44} & \cellcolor{rk050}7.44 & \cellcolor{rk050}9.19 & \cellcolor{rk000}\best{7.62} & \cellcolor{rk050}5.33 & \cellcolor{rk100}6.12 & \cellcolor{rk050}15.8 \\
\rowcolor{oursrow} & \method{} & \cellcolor{rk000}\best{14.0} & \cellcolor{rk000}\best{12.1} & \cellcolor{rk100}16.9 & \cellcolor{rk000}\best{21.5} & \cellcolor{rk000}\best{18.3} & \cellcolor{rk050}8.58 & \cellcolor{rk000}\best{6.18} & \cellcolor{rk000}\best{8.56} & \cellcolor{rk100}9.51 & \cellcolor{rk100}5.99 & \cellcolor{rk000}\best{3.30} & \cellcolor{rk000}\best{11.4} \\
\cmidrule(l){2-14}
 & Identity & \cellcolor{rk100}60.7 & \cellcolor{rk100}67.4 & \cellcolor{rk050}25.0 & \cellcolor{rk100}59.4 & \cellcolor{rk100}44.3 & \cellcolor{rk050}18.2 & \cellcolor{rk000}\best{19.0} & \cellcolor{rk050}22.2 & \cellcolor{rk050}16.2 & \cellcolor{rk050}8.73 & \cellcolor{rk050}19.3 & \cellcolor{rk100}32.8 \\
VGGT & Heuristic & \cellcolor{rk050}44.2 & \cellcolor{rk050}46.8 & \cellcolor{rk050}25.0 & \cellcolor{rk050}54.2 & \cellcolor{rk050}42.9 & \cellcolor{rk100}20.0 & \cellcolor{rk100}21.0 & \cellcolor{rk100}23.8 & \cellcolor{rk100}16.8 & \cellcolor{rk100}10.5 & \cellcolor{rk100}22.6 & \cellcolor{rk050}29.8 \\
\bottomrule
\end{tabular}

\end{table*}
}
\suppfull{\begin{table*}[t]
\centering
\footnotesize
\setlength{\tabcolsep}{3.1pt}
\caption{\textbf{Per-sequence visual--inertial initialization on Aria.} Settings follow \cref{sec:supp_downstream}; aggregation follows \cref{sec:supp_perseq}.}
\label{tab:supp_aria_vi}
\begin{tabular}{@{}llcccccccccc@{}}
\toprule
Backbone & Weighting & AprilTag1 & AprilTag2 & BigRoom2 & BigRoom3 & BigRoom4 & NewTags1 & NewTags2 & NewTags3 & NewTags4 & Mean \\
\midrule
\multicolumn{12}{@{}l}{\textit{$e_s$ (\%)}} \\
\cmidrule(r){1-2}
 & Identity & \cellcolor{rk100}13.3 & \cellcolor{rk050}13.9 & \cellcolor{rk000}\best{12.4} & \cellcolor{rk050}10.3 & \cellcolor{rk050}11.2 & \cellcolor{rk050}13.1 & \cellcolor{rk050}14.7 & \cellcolor{rk050}9.24 & \cellcolor{rk050}11.3 & \cellcolor{rk050}12.2 \\
VGGT-$\Omega$ & Heuristic & \cellcolor{rk050}13.2 & \cellcolor{rk100}14.0 & \cellcolor{rk000}\best{12.4} & \cellcolor{rk050}10.3 & \cellcolor{rk100}11.4 & \cellcolor{rk100}13.5 & \cellcolor{rk050}14.7 & \cellcolor{rk100}9.35 & \cellcolor{rk100}11.5 & \cellcolor{rk100}12.3 \\
\rowcolor{oursrow} & \method{} & \cellcolor{rk000}\best{9.74} & \cellcolor{rk000}\best{11.0} & \cellcolor{rk100}14.3 & \cellcolor{rk000}\best{9.46} & \cellcolor{rk000}\best{10.1} & \cellcolor{rk000}\best{9.66} & \cellcolor{rk000}\best{10.2} & \cellcolor{rk000}\best{7.76} & \cellcolor{rk000}\best{8.24} & \cellcolor{rk000}\best{10.0} \\
\cmidrule(l){2-12}
 & Identity & \cellcolor{rk100}32.0 & \cellcolor{rk100}35.2 & \cellcolor{rk050}22.8 & \cellcolor{rk050}20.5 & \cellcolor{rk050}30.2 & \cellcolor{rk050}32.1 & \cellcolor{rk050}41.4 & \cellcolor{rk050}33.0 & \cellcolor{rk050}33.6 & \cellcolor{rk050}31.2 \\
VGGT & Heuristic & \cellcolor{rk050}31.9 & \cellcolor{rk050}34.9 & \cellcolor{rk000}\best{22.6} & \cellcolor{rk100}20.6 & \cellcolor{rk100}30.3 & \cellcolor{rk050}32.1 & \cellcolor{rk100}41.5 & \cellcolor{rk050}33.0 & \cellcolor{rk100}33.7 & \cellcolor{rk050}31.2 \\
\rowcolor{oursrow} & \method{} & \cellcolor{rk000}\best{23.3} & \cellcolor{rk000}\best{24.8} & \cellcolor{rk100}24.1 & \cellcolor{rk000}\best{16.2} & \cellcolor{rk000}\best{25.4} & \cellcolor{rk000}\best{22.7} & \cellcolor{rk000}\best{27.7} & \cellcolor{rk000}\best{23.2} & \cellcolor{rk000}\best{21.8} & \cellcolor{rk000}\best{23.3} \\
\bottomrule
\end{tabular}

\end{table*}
}
\supppair{\begin{table*}[t]
\centering
\footnotesize
\setlength{\tabcolsep}{3.2pt}
\caption{\textbf{Per-sequence visual--inertial initialization on ScaRF.} Settings follow \cref{sec:supp_downstream}; aggregation follows \cref{sec:supp_perseq}.}
\label{tab:supp_scarf_vi}
\begin{tabular}{@{}llcccccc@{}}
\toprule
Backbone & Weighting & r01 & r02 & r03 & r04 & r05 & Mean \\
\midrule
\multicolumn{8}{@{}l}{\textit{$e_s$ (\%)}} \\
\cmidrule(r){1-2}
 & Identity & \cellcolor{rk100}10.7 & \cellcolor{rk100}9.78 & \cellcolor{rk050}11.4 & \cellcolor{rk100}8.05 & \cellcolor{rk100}7.92 & \cellcolor{rk100}9.56 \\
VGGT-$\Omega$ & Heuristic & \cellcolor{rk050}10.3 & \cellcolor{rk050}9.44 & \cellcolor{rk050}11.4 & \cellcolor{rk050}7.78 & \cellcolor{rk050}7.47 & \cellcolor{rk050}9.28 \\
\rowcolor{oursrow} & \method{} & \cellcolor{rk000}\best{7.24} & \cellcolor{rk000}\best{5.81} & \cellcolor{rk000}\best{6.62} & \cellcolor{rk000}\best{4.91} & \cellcolor{rk000}\best{6.29} & \cellcolor{rk000}\best{6.17} \\
\cmidrule(l){2-8}
 & Identity & \cellcolor{rk100}27.6 & \cellcolor{rk100}35.7 & \cellcolor{rk100}31.7 & \cellcolor{rk100}26.2 & \cellcolor{rk100}30.4 & \cellcolor{rk100}30.3 \\
VGGT & Heuristic & \cellcolor{rk050}25.9 & \cellcolor{rk050}34.0 & \cellcolor{rk050}29.9 & \cellcolor{rk050}24.4 & \cellcolor{rk050}28.3 & \cellcolor{rk050}28.5 \\
\rowcolor{oursrow} & \method{} & \cellcolor{rk000}\best{18.2} & \cellcolor{rk000}\best{20.4} & \cellcolor{rk000}\best{17.2} & \cellcolor{rk000}\best{11.9} & \cellcolor{rk000}\best{17.6} & \cellcolor{rk000}\best{17.1} \\
\bottomrule
\end{tabular}

\end{table*}
}{\begin{table*}[t]
\centering
\footnotesize
\setlength{\tabcolsep}{6pt}
\caption{\textbf{Per-sequence visual--inertial initialization on Spires.} Settings follow \cref{sec:supp_downstream}; aggregation follows \cref{sec:supp_perseq}.}
\label{tab:supp_spires_vi}
\begin{tabular}{@{}llcccc@{}}
\toprule
Backbone & Weighting & CC02 & CC03 & CC05 & Mean \\
\midrule
\multicolumn{6}{@{}l}{\textit{$e_s$ (\%)}} \\
\cmidrule(r){1-2}
 & Identity & \cellcolor{rk050}34.8 & \cellcolor{rk100}28.6 & \cellcolor{rk050}32.2 & \cellcolor{rk050}31.9 \\
VGGT-$\Omega$ & Heuristic & \cellcolor{rk100}35.0 & \cellcolor{rk000}\best{28.4} & \cellcolor{rk100}32.5 & \cellcolor{rk100}32.0 \\
\rowcolor{oursrow} & \method{} & \cellcolor{rk000}\best{33.1} & \cellcolor{rk000}\best{28.4} & \cellcolor{rk000}\best{31.3} & \cellcolor{rk000}\best{30.9} \\
\cmidrule(l){2-6}
 & Identity & \cellcolor{rk050}46.1 & \cellcolor{rk100}42.7 & \cellcolor{rk100}42.3 & \cellcolor{rk100}43.7 \\
VGGT & Heuristic & \cellcolor{rk050}46.1 & \cellcolor{rk050}42.0 & \cellcolor{rk050}42.2 & \cellcolor{rk050}43.4 \\
\rowcolor{oursrow} & \method{} & \cellcolor{rk000}\best{41.7} & \cellcolor{rk000}\best{38.3} & \cellcolor{rk000}\best{38.9} & \cellcolor{rk000}\best{39.6} \\
\bottomrule
\end{tabular}

\end{table*}
}
\suppfull{\begin{table*}[t]
\centering
\footnotesize
\setlength{\tabcolsep}{2.6pt}
\caption{\textbf{Per-sequence SLAM results on EuRoC.} Settings follow \cref{sec:supp_downstream}; aggregation follows \cref{sec:supp_perseq}. \fail{} indicates divergence (translational RPE above 5\,m/m).}
\label{tab:supp_euroc_systems}
\begin{tabular}{@{}llcccccccccccc@{}}
\toprule
Backbone & Weighting & MH01 & MH02 & MH03 & MH04 & MH05 & V101 & V102 & V103 & V201 & V202 & V203 & Mean \\
\midrule
\multicolumn{14}{@{}l}{\textit{VGGT-Bridge without loop closure, Sim(3) ATE RMSE (m)}} \\
\cmidrule(r){1-2}
VGGT-$\Omega$ & Default & \cellcolor{rk100}0.196 & \cellcolor{rk000}\best{0.134} & \cellcolor{rk100}0.253 & \cellcolor{rk100}0.682 & \cellcolor{rk100}0.543 & \cellcolor{rk100}0.071 & \cellcolor{rk000}\best{0.055} & \cellcolor{rk100}0.084 & \cellcolor{rk100}0.097 & \cellcolor{rk100}0.055 & \cellcolor{rk100}0.073 & \cellcolor{rk100}0.204 \\
\rowcolor{oursrow} & \method{} & \cellcolor{rk000}\best{0.131} & \cellcolor{rk100}0.150 & \cellcolor{rk000}\best{0.187} & \cellcolor{rk000}\best{0.238} & \cellcolor{rk000}\best{0.468} & \cellcolor{rk000}\best{0.061} & \cellcolor{rk100}0.064 & \cellcolor{rk000}\best{0.069} & \cellcolor{rk000}\best{0.069} & \cellcolor{rk000}\best{0.044} & \cellcolor{rk000}\best{0.060} & \cellcolor{rk000}\best{0.140} \\
\cmidrule(l){2-14}
VGGT & Default & \cellcolor{rk100}0.347 & \cellcolor{rk100}0.271 & \cellcolor{rk100}0.572 & \cellcolor{rk100}3.908 & \cellcolor{rk100}6.034 & \cellcolor{rk000}\best{0.105} & \cellcolor{rk100}0.357 & \cellcolor{rk100}0.156 & \cellcolor{rk100}0.086 & \cellcolor{rk100}0.075 & \cellcolor{rk100}0.180 & \cellcolor{rk100}1.099 \\
\rowcolor{oursrow} & \method{} & \cellcolor{rk000}\best{0.176} & \cellcolor{rk000}\best{0.110} & \cellcolor{rk000}\best{0.241} & \cellcolor{rk000}\best{2.984} & \cellcolor{rk000}\best{5.902} & \cellcolor{rk000}\best{0.105} & \cellcolor{rk000}\best{0.146} & \cellcolor{rk000}\best{0.107} & \cellcolor{rk000}\best{0.075} & \cellcolor{rk000}\best{0.058} & \cellcolor{rk000}\best{0.129} & \cellcolor{rk000}\best{0.912} \\
\bottomrule
\end{tabular}

\end{table*}
}
\suppfull{\begin{table*}[t]
\centering
\footnotesize
\setlength{\tabcolsep}{3.1pt}
\caption{\textbf{Per-sequence SLAM results on Aria.} Settings follow \cref{sec:supp_downstream}; aggregation follows \cref{sec:supp_perseq}. \fail{} indicates divergence (translational RPE above 5\,m/m).}
\label{tab:supp_aria_systems}
\begin{tabular}{@{}llcccccccccc@{}}
\toprule
Backbone & Weighting & AprilTag1 & AprilTag2 & BigRoom2 & BigRoom3 & BigRoom4 & NewTags1 & NewTags2 & NewTags3 & NewTags4 & Mean \\
\midrule
\multicolumn{12}{@{}l}{\textit{VGGT-Bridge without loop closure, Sim(3) ATE RMSE (m)}} \\
\cmidrule(r){1-2}
VGGT-$\Omega$ & Default & \cellcolor{rk000}\best{0.681} & \cellcolor{rk100}0.373 & \cellcolor{rk100}0.085 & \cellcolor{rk000}\best{0.064} & \cellcolor{rk000}\best{0.057} & \cellcolor{rk000}\best{0.241} & \cellcolor{rk100}1.285 & \cellcolor{rk100}1.614 & \cellcolor{rk100}1.219 & \cellcolor{rk100}0.624 \\
\rowcolor{oursrow} & \method{} & \cellcolor{rk100}0.743 & \cellcolor{rk000}\best{0.370} & \cellcolor{rk000}\best{0.079} & \cellcolor{rk100}0.079 & \cellcolor{rk100}0.058 & \cellcolor{rk100}0.272 & \cellcolor{rk000}\best{0.403} & \cellcolor{rk000}\best{0.768} & \cellcolor{rk000}\best{0.535} & \cellcolor{rk000}\best{0.368} \\
\cmidrule(l){2-12}
VGGT & Default & \cellcolor{rk000}\best{3.716} & \cellcolor{rk100}6.436 & \cellcolor{rk100}0.393 & \cellcolor{rk100}0.897 & \cellcolor{rk000}\best{0.496} & \cellcolor{rk100}4.754 & \cellcolor{rk100}4.883 & \cellcolor{rk100}7.227 & \cellcolor{rk100}6.322 & \cellcolor{rk100}3.903 \\
\rowcolor{oursrow} & \method{} & \cellcolor{rk100}4.731 & \cellcolor{rk000}\best{5.868} & \cellcolor{rk000}\best{0.317} & \cellcolor{rk000}\best{0.619} & \cellcolor{rk100}0.551 & \cellcolor{rk000}\best{4.232} & \cellcolor{rk000}\best{4.702} & \cellcolor{rk000}\best{6.533} & \cellcolor{rk000}\best{5.541} & \cellcolor{rk000}\best{3.677} \\
\cmidrule(l){2-12}
DA3 & Default & \cellcolor{rk100}1.570 & \cellcolor{rk100}0.990 & \cellcolor{rk000}\best{0.096} & \cellcolor{rk000}\best{0.093} & \cellcolor{rk000}\best{0.104} & \cellcolor{rk100}0.553 & \cellcolor{rk100}1.993 & \cellcolor{rk000}\best{4.893} & \cellcolor{rk100}5.669 & \cellcolor{rk100}1.773 \\
\bottomrule
\end{tabular}

\end{table*}
}
\supppair{\begin{table*}[t]
\centering
\footnotesize
\setlength{\tabcolsep}{3.2pt}
\caption{\textbf{Per-sequence SLAM results on ScaRF.} Settings follow \cref{sec:supp_downstream}; aggregation follows \cref{sec:supp_perseq}. \fail{} indicates divergence (translational RPE above 5\,m/m).}
\label{tab:supp_scarf_systems}
\begin{tabular}{@{}llcccccc@{}}
\toprule
Backbone & Weighting & r01 & r02 & r03 & r04 & r05 & Mean \\
\midrule
\multicolumn{8}{@{}l}{\textit{VGGT-Bridge without loop closure, Sim(3) ATE RMSE (m)}} \\
\cmidrule(r){1-2}
VGGT-$\Omega$ & Default & \cellcolor{rk000}\best{0.484} & \cellcolor{rk100}2.795 & \cellcolor{rk000}\best{3.457} & \cellcolor{rk100}0.654 & \cellcolor{rk100}0.339 & \cellcolor{rk100}1.546 \\
\rowcolor{oursrow} & \method{} & \cellcolor{rk100}0.540 & \cellcolor{rk000}\best{0.886} & \cellcolor{rk100}3.814 & \cellcolor{rk000}\best{0.202} & \cellcolor{rk000}\best{0.335} & \cellcolor{rk000}\best{1.155} \\
\cmidrule(l){2-8}
VGGT & Default & \cellcolor{rk100}5.693 & \cellcolor{rk000}\best{3.534} & \cellcolor{rk100}3.578 & \cellcolor{rk000}\best{3.975} & \cellcolor{rk100}3.907 & \cellcolor{rk100}4.137 \\
\rowcolor{oursrow} & \method{} & \cellcolor{rk000}\best{4.667} & \cellcolor{rk100}3.621 & \cellcolor{rk000}\best{3.424} & \cellcolor{rk100}4.369 & \cellcolor{rk000}\best{1.954} & \cellcolor{rk000}\best{3.607} \\
\cmidrule(l){2-8}
DA3 & Default & \cellcolor{rk000}\best{2.830} & \cellcolor{rk100}2.866 & \cellcolor{rk100}1.323 & \cellcolor{rk100}1.287 & \cellcolor{rk100}0.943 & \cellcolor{rk100}1.850 \\
\rowcolor{oursrow} & \method{} & \cellcolor{rk100}3.117 & \cellcolor{rk000}\best{2.364} & \cellcolor{rk000}\best{0.654} & \cellcolor{rk000}\best{0.602} & \cellcolor{rk000}\best{0.354} & \cellcolor{rk000}\best{1.418} \\
\bottomrule
\end{tabular}

\end{table*}
}{\begin{table*}[t]
\centering
\footnotesize
\setlength{\tabcolsep}{6pt}
\caption{\textbf{Per-sequence SLAM results on Spires.} Settings follow \cref{sec:supp_downstream}; aggregation follows \cref{sec:supp_perseq}. \fail{} indicates divergence (translational RPE above 5\,m/m).}
\label{tab:supp_spires_systems}
\begin{tabular}{@{}llcccc@{}}
\toprule
Backbone & Weighting & CC02 & CC03 & CC05 & Mean \\
\midrule
\multicolumn{6}{@{}l}{\textit{VGGT-Bridge without loop closure, Sim(3) ATE RMSE (m)}} \\
\cmidrule(r){1-2}
VGGT-$\Omega$ & Default & \cellcolor{rk000}\best{3.744} & \cellcolor{rk000}\best{1.092} & \cellcolor{rk000}\best{6.512} & \cellcolor{rk000}\best{3.783} \\
\rowcolor{oursrow} & \method{} & \cellcolor{rk100}4.926 & \cellcolor{rk100}1.157 & \cellcolor{rk100}7.243 & \cellcolor{rk100}4.442 \\
\cmidrule(l){2-6}
VGGT & Default & \cellcolor{rk100}20.20 & \cellcolor{rk000}\best{15.76} & \cellcolor{rk100}17.36 & \cellcolor{rk100}17.77 \\
\rowcolor{oursrow} & \method{} & \cellcolor{rk000}\best{8.053} & \cellcolor{rk100}16.14 & \cellcolor{rk000}\best{9.653} & \cellcolor{rk000}\best{11.28} \\
\cmidrule(l){2-6}
DA3 & Default & \cellcolor{rk100}4.225 & \cellcolor{rk100}2.860 & \cellcolor{rk100}6.744 & \cellcolor{rk100}4.610 \\
\rowcolor{oursrow} & \method{} & \cellcolor{rk000}\best{2.347} & \cellcolor{rk000}\best{2.677} & \cellcolor{rk000}\best{5.822} & \cellcolor{rk000}\best{3.615} \\
\bottomrule
\end{tabular}

\end{table*}
}
\endgroup

\end{document}